%% file: acl_latex.tex
\pdfoutput=1
\documentclass[11pt]{article}

\usepackage[preprint]{acl}

\usepackage{amsmath}
\usepackage{adjustbox}
\usepackage{amssymb}    % For additional mathematical symbols
\usepackage{amsthm}     % For theorem, lemma, and proof environments
\usepackage{tcolorbox} % For creating colored boxes (like the one used for the prompt)
\usepackage{xcolor}    % For color management used by tcolorbox
\usepackage{latexsym}
\usepackage{times}
\usepackage{algorithm}    % For the algorithm environment
\usepackage[noend]{algpseudocode} % For pseudocode commands like \State, \For, etc.algorithmic environment
\usepackage{float}        % To allow the [H] specifier for precise placement

\usepackage[T1]{fontenc}
\usepackage[utf8]{inputenc}

\usepackage{microtype}

\usepackage{inconsolata}

\usepackage{graphicx}

\usepackage{hyperref}
\usepackage{float}
\usepackage{booktabs}
\usepackage{multirow}
\usepackage{wrapfig}
\usepackage{natbib}
\usepackage{microtype}
\usepackage{xcolor}         % colors
\usepackage{amsmath}
\usepackage{amssymb}
\usepackage{multicol}
\usepackage{wrapfig}
\usepackage{tabularx}
\usepackage{booktabs}       % professional-quality tables
\usepackage{amsfonts}       % blackboard math symbols
\usepackage{graphicx}
\usepackage{nicefrac}  
\usepackage{pifont}
\usepackage{colortbl}
\usepackage{longtable}
\usepackage{ulem}
\usepackage{listings}
\usepackage{xspace}
\usepackage{fancyvrb}
\usepackage{adjustbox}
\usepackage{amsthm}
\usepackage{tikz}
\usepackage{forest}
\usepackage{csquotes}

\newcommand{\ourmodel}[0]{{\textsc{PromptCoT}}\xspace}

\newtheorem{lemma}{Lemma}

\title{\ourmodel: Synthesizing Olympiad-level Problems for Mathematical Reasoning in Large Language Models}

\author{
\textbf{Xueliang Zhao}$^\spadesuit$\thanks{\hspace{2mm}This work was done during an internship at Ant Group.} \quad
\textbf{Wei Wu}$^{\bigstar}$\thanks{\hspace{2mm}Corresponding authors.} \quad
\textbf{Jian Guan}$^{\bigstar}$ \quad
\textbf{Lingpeng Kong}$^\spadesuit$\footnotemark[2] \\
  $^\spadesuit$The University of Hong Kong \quad
  $^\bigstar$Ant Group \\
\texttt{\{xlzhao,lpk\}@cs.hku.hk}\\
\texttt{\{wuwei19850318, jianguanthu\}@gmail.com} \\
}

\begin{document}
\maketitle
\input{0-abstract}

\input{1-introduction}

\input{2-method}

\input{3-experiments}

\input{4-related}
\input{5-conclusion}

\bibliography{custom}

\input{6-appendix}

\end{document}

%% file: 0-abstract.tex
\begin{abstract}
The ability of large language models to solve complex mathematical problems has progressed significantly, particularly for tasks requiring advanced reasoning. However, the scarcity of sufficiently challenging problems, particularly at the Olympiad level, hinders further advancements. In this work, we introduce \textsc{PromptCoT}, a novel approach for automatically generating high-quality Olympiad-level math problems. The proposed method synthesizes complex problems based on mathematical concepts and the rationale behind problem construction, emulating the thought processes of experienced problem designers. We provide a theoretical analysis demonstrating that an optimal rationale should maximize both the likelihood of rationale generation given the associated concepts and the likelihood of problem generation conditioned on both the rationale and the concepts. Our method is evaluated on standard benchmarks including GSM8K, MATH-500, and AIME2024, where it consistently outperforms existing problem generation methods. Furthermore, we demonstrate that \textsc{PromptCoT} exhibits superior data scalability, consistently maintaining high performance as the dataset size increases, outperforming the baselines. The implementation is available at \url{https://github.com/zhaoxlpku/PromptCoT}.
\end{abstract}

%% file: 1-introduction.tex
\begin{table*}[!ht]
\centering
\begin{adjustbox}{max width=\linewidth}
\begin{tabular}{@{}lccccc@{}}
\toprule
\multirow{2}{*}{\textbf{Dataset}} & \textbf{Qwen2.5-Math-72B-Instruct}&\textbf{DeepSeek-R1-Distill-Qwen-7B}& \multicolumn{3}{c}{\textbf{Micro Avg. Accuracy on MATH-500 and AIME2024}} \\
\cline{4-6}
&\textbf{Accuracy ($\downarrow$)} & \textbf{Avg. Reasoning Tokens ($\uparrow$)} & \textbf{Fine-tuned Qwen2.5-Math-7B} &\textbf{$\Delta$ ($\uparrow$)} \\
\midrule
\textbf{AIME2024} & \textit{30.0} & \textit{4,159} & - &-\\ \midrule
\textbf{\ourmodel} (ours) & \textbf{48.9} & \textbf{6,502} & 80.8  &\textcolor{red}{\textbf{+1.2}} \\
\textbf{OpenMathInstruct}~\citep{toshniwal2024openmathinstruct} & 63.3 & 1,578 & 75.8 & \textcolor{green}{-3.8} \\
\textbf{NuminaMath}~\citep{li2024numinamath} & 60.9 & 3,039 & 76.6  &\textcolor{green}{-3.0} \\
\textbf{Evol-Instruct}~\citep{luo2023wizardmath} & 65.9 & 1,346 & 74.0  &\textcolor{green}{-5.6} \\
\textbf{KPDDS}~\citep{huang2024key} & 73.2 & 1,225 &72.3 &\textcolor{green}{-7.3} \\
\bottomrule
\end{tabular}
\end{adjustbox}
\caption{
Difficulty and efficacy evaluation for different mathematical datasets. \textbf{Accuracy:} Performance of Qwen2.5-Math-72B-Instruct on the problems in different datasets. Note that we directly apply the model to solve the problems without any tuning, so the metric reflects difficulty of the problems for Qwen2.5-Math-72B-Instruct ($\downarrow$: lower accuracy indicates higher difficulty). \textbf{Avg. Reasoning Tokens:} Average number of tokens in reasoning processes generated by DeepSeek-R1-Distill-Qwen-7B when processing the problems in different datasets ($\uparrow$: larger numbers means DeepSeek-R1-Distill-Qwen-7B needs more tokens to complete reasoning, suggesting greater problem difficulty). \textbf{Micro Avg. Accuracy on MATH-500 and AIME2024:} Performance of Qwen2.5-Math-7B after fine-tuning on different datasets. Accuracy is computed as a weighted average over MATH-500 and AIME 2024. Note that fine-tuning is conducted using both the problems and their corresponding solutions. Further details are provided in Section \ref{sec:exp}. \textbf{$\Delta$:} Difference in performance between Qwen2.5-Math-7B and Qwen2.5-Math-7B-Instruct ($\uparrow$: larger margins suggests bigger contributions from the corresponding datasets). Note that no tuning is performed on Qwen2.5-Math-7B-Instruct, so its Micro Avg. Accuracy remains fixed at $79.6$.}
\label{tab:introduction}
\end{table*}

\section{Introduction}\label{sec:intro}

\begin{displayquote}
\textit{``In mathematics the art of proposing a question must be held of higher value than solving it.''}
\begin{flushright}
\footnotesize ------ Georg Cantor
\end{flushright}
\end{displayquote}

Recent advancements in large language models (LLMs) have greatly enhanced their capability for solving complex problems through planning and reasoning. Particularly in mathematics, strong reasoning models such as OpenAI o1 \cite{jaech2024openai} and DeepSeek r1 \cite{guo2025deepseek} have significantly pushed the boundaries of AI from mastering grade school problems \cite{yang2024qwen2} to excelling at Olympiad-level challenges \cite{guo2025deepseek}. 
The remarkable achievements have inspired the community to explore dedicating more computational resources to the inference stage. Consequently, the scaling paradigm of LLMs is shifting from training time to test time \cite{snell2024scaling}. 
While powerful reasoning LLMs, such as DeepSeek r1, have been open-sourced, the details of math problem acquisition remain obscured in the published literature (e.g., the technical report of DeepSeek r1), and such data is still kept private. Consequently, studies aiming to reproduce o1-like or r1-like performance have to rely on open-source math datasets \cite{li2024numinamath}, which are often constrained by scale and difficulty. This leaves an open research question (\textit{RQ}): \textit{how can we obtain high-quality, sufficiently challenging math problems at scale?} We emphasize that \textit{RQ} is crucial for test-time scaling research, as it serves as a prerequisite for effectively initiating the supervised fine-tuning (SFT) or reinforcement learning (RL) process. 

We investigate the automated generation of Olympiad-level math problems as a principled approach to addressing \textit{RQ}. Prior to our work, several projects have curated math datasets, typically involving problem synthesis procedures. Existing synthesis methods can be categorized into three frameworks: (1) direct generation through prompting powerful LLMs \cite{huang2024key, tang2024mathscale, li2024numinamath, toshniwal2024openmathinstruct}, (2) mining from pretraining corpora \cite{yue2023mammothbuildingmathgeneralist, li2024selfalignmentinstructionbacktranslation}, and (3) problem evolution, either through simple-to-simple rephrasing \cite{yu2023metamath} or difficulty-based progression \cite{xu2023wizardlm, luo2023wizardmath}.
While these efforts progressively contribute to LLM reasoning, the problems provided by these methods now are not challenging enough for state-of-the-art models, making them less effective in further advancements. Table \ref{tab:introduction} provides a detailed analysis of problem difficulty in typical published works and open-source datasets, along with their efficacy in enhancing state-of-the-art LLMs.
From this, we observe clear gaps in terms of difficulty between the problems in existing work and those in AIME, as well as the limited utility of these problems in further improving LLM performance.

\begin{figure*}[!t]
\centering
\includegraphics[width=0.8\textwidth]{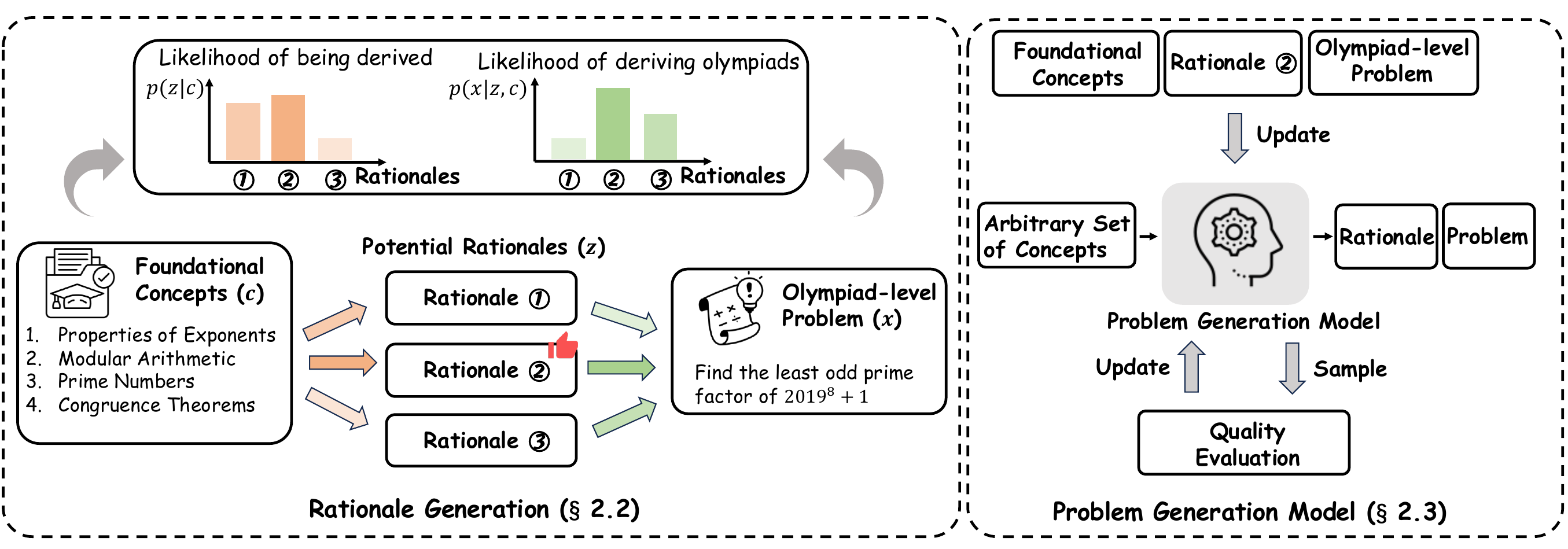}
\caption{Overview of the proposed method. \textbf{Left:} Given an Olympiad problem and its foundational concepts (extracted using an LLM), the goal is to identify rationales that act as ``amplifiers'' to enhance the likelihood of generating the Olympiad problem (i.e., $p(x \mid z, \mathbf{c})$) while ensuring that these rationales can be easily derived from the foundational concepts (i.e.,  $p(z \mid \mathbf{c})$). \textbf{Right:} Once an appropriate rationale is found, we use the \(\langle \text{foundational concepts}, \text{rationale}, \text{Olympiad problem} \rangle\) triple to train a model capable of predicting both the rationale and the Olympiad problem from any given set of concepts. The model can be further optimized through rejection sampling, where the quality of generated outputs is evaluated and used for subsequent model training.} \label{fig:method_overview} 
 \end{figure*}

The primary challenge in synthesizing Olympiad-level math problems lies in their scarcity in existing corpora. As a result, the complex reasoning patterns required for such problems are undertrained, making it difficult for LLMs to assign sufficient probability to them during inference. More broadly, what we aim to explore is by nature a low-resource generation problem, which is prevalent across various applications yet inherently contrasts with the fundamental working mechanisms of LLMs. We focus on the mathematical domain, but our method can be easily adapted to other domains. Specifically, we propose \ourmodel, a novel problem generation method that synthesizes a complex math problem based on given math concepts (e.g., ``Prime Numbers'', c.f. Figure~\ref{fig:method_overview}) and a rationale that emulates a veteran teacher's thought process when designing math problems for students. The idea is inspired by the success of LLMs in ``solving'' difficult problems, where detailed reasoning procedures (e.g., chain-of-thoughts \citep{wei2022chain}) have proven particularly effective. Through theoretical analysis, we show that an optimal rationale should simultaneously maximize the likelihood of rationale generation given the concepts and the likelihood of problem generation conditioned on both the rationale and the concepts. Based on this analysis, we implement \ourmodel by first querying an LLM to generate rationales for prepared Olympiad-level math problems and their associated concepts, and then fine-tuning an LLM as a problem generation model using a set of concept-rationale-problem triples.

We conduct extensive experiments on multiple benchmarks focused on mathematical reasoning, including GSM8K~\citep{cobbe2021training}, MATH-500~\citep{lightman2023let}, and AIME2024~\citep{aimedata}. Evaluation results demonstrate that, compared to a range of existing generation methods and datasets, \ourmodel leads to more significant improvements on state-of-the-art LLMs after distillation in both short-CoT and long-CoT settings, achieving \textbf{0.4\%-4.8\%} absolute gains on MATH-500 and \textbf{6.7\%-20\%} absolute gains on AIME2024. An extended experiment further shows that \ourmodel exhibits remarkable advantages over the baseline method as the number of problems gradually increases, highlighting its superior data scalability.

Our contributions are three-fold: (1) We propose synthesizing Olympiad-level math problems to enhance LLM reasoning. While previous research \cite{snell2024scaling} has identified problem difficulty as a key factor in improving LLMs’ reasoning capabilities, to the best of our knowledge, we are the first to formally pose difficult problem generation as a research question (i.e., \textit{RQ}); (2) We propose \ourmodel as a principled approach to generating Olympiad-level math problems. To the best of our knowledge, we are the first to apply the chain-of-thought paradigm to the task of problem generation; and (3) We conduct extensive experiments to evaluate the efficacy of \ourmodel. Notably, \ourmodel is the only method that enables Qwen2.5-Math base models to surpass their Instruct versions. Furthermore, a 7B model distilled using the problems generated by \ourmodel is able to achieve performance comparable to state-of-the-art 32B models on mathematical reasoning tasks.

%% file: 2-method.tex
\begin{algorithm*}
\small
\caption{Rationale-Guided Problem Generation}
\label{alg:prompt_generation}
\begin{algorithmic}[1]
\Require A set of seed prompts \(\{p_1, p_2, \ldots, p_n\}\) and an LLM for concept extraction and rationale generation.
\State Initialize training set \(T \gets \emptyset\).
\ForAll{seed prompt \(x \in \{x_1, x_2, \ldots, x_n\}\)}
    \State \textbf{Concept Extraction:} Query the LLM (using the instruction in Appendix~\ref{sec:appendix_concept}) to extract a set of foundational concepts \(\mathbf{c}\).
    \State \textbf{Rationale Generation:} Query the LLM (using the instruction in Appendix~\ref{sec:appendix_rationale}) with \(x\) and \(\mathbf{c}\) to generate a rationale \(z\).
    \State Add the triplet \(\langle \mathbf{c}, z, x \rangle\) to \(T\).
\EndFor
\State \textbf{Model Training:} Train a  problem generation model on \(T\) with MLE and rejection sampling.
\State \textbf{Inference:} For any given set of foundational concepts \(\mathbf{c}'\), use the trained model to generate a rationale \(z'\) and a problem \(x'\), and aggregate these problems into the set \(\{x'_1, x'_2, \ldots, x'_m\}\).
\Statex \hspace*{-\algorithmicindent} \textbf{Output:} A set of high-quality and challenging problems \(\{x'_1, x'_2, \ldots, x'_m\}\).
\end{algorithmic}
\end{algorithm*}

\section{Method}
Figure \ref{fig:method_overview} provides an overview of \ourmodel. In summary, our method comprises three key steps: (1) \textbf{Concept Extraction:} Foundational concepts are extracted from seed prompts collected from Olympiad math sources~(\S\ref{sec:concept}); (2) \textbf{Rationale Generation:} A rationale is inferred for each problem based on its associated concepts by maximizing both the probability of the rationale given the concepts and the probability of the problem given both the rationale and the concepts~(\S\ref{sec:rationale}); and (3) \textbf{Problem Generation Model Training:} Concept-rationale-problem triples are used to fine-tune an LLM as a problem generation model, followed by a rejection sampling fine-tuning stage for further self-improvement, ensuring the quality of generated problems~(\S\ref{sec:training}). 

\subsection{Concept Extraction}\label{sec:concept}
We collect a large set of seed prompts from the AoPS~\footnote{\url{https://artofproblemsolving.com/}} platform which contains challenging Olympiad-level math problems. 
Problems overlapping with common test benchmarks are filtered out to prevent data leakage. For each seed prompt, domain-specific concepts are extracted using a large language model. The model is provided with a query instruction (see Appendix~\ref{sec:appendix_concept} for details) that directs it to identify the salient concepts associated with the problem. While tailored to mathematics in our study, these concepts serve as a concise abstraction of the underlying reasoning in problem design and can be readily adapted to other domains by replacing the seed prompts with those relevant to the target domain.

\subsection{Rationale Generation}\label{sec:rationale}

For effective problem generation, the underlying rationale should (i) be naturally derived from a set of foundational concepts and (ii) increase the likelihood of producing a high-quality problem. To capture this idea, let us denote the set of extracted $K$ concepts by
\[
\mathbf{c} = \{c_1, c_2, \ldots, c_K\}.
\]
We introduce a latent variable \(z\) that represents the rationale connecting these concepts to the original problem $x$. The generation process is then formulated as
\[
p(x \mid \mathbf{c}) = \sum_{z} p(x, z \mid \mathbf{c})= \sum_{z} p(x \mid z, \mathbf{c}) \, p(z \mid \mathbf{c}).
\]
Ideally, we aim to maximize \(p(x \mid \mathbf{c})\); however, marginalizing over the latent variable $z$ is intractable. Therefore, we introduce a variational distribution \(q(z \mid \mathbf{c}, x)\) to approximate the true posterior \(p(z \mid \mathbf{c}, x)\) and derive an evidence lower bound (ELBO) via Jensen’s inequality:
\[
\log p(x \mid \mathbf{c}) \ge \mathbb{E}_{q(z \mid \mathbf{c}, x)} \left[ \log \frac{p(x, z \mid \mathbf{c})}{q(z \mid \mathbf{c}, x)} \right].
\]

\begin{lemma}[Optimal Variational Distribution]
The optimal variational distribution \(q^\star(z \mid \mathbf{c}, x)\) that maximizes the ELBO satisfies
\[
q^\star(z \mid \mathbf{c}, x) \propto p(x \mid z, \mathbf{c}) p(z \mid \mathbf{c}).
\]
\end{lemma}
\noindent The proof is provided in Appendix~\ref{sec:appendix_proof}. 
This result implies that the optimal distribution over the latent rationale is governed by two key factors: the extent to which the rationale is naturally derived from the set of foundational concepts \(\mathbf{c}\) (as indicated by \(p(z \mid \mathbf{c})\)), and the degree to which this rationale increases the likelihood of generating a high-quality problem (as indicated by \(p(x \mid z, \mathbf{c})\)). In our framework, these insights ensure that the inferred rationale not only reflects the core input information but also contributes to the production of the problems.
In practice, \(z\) is obtained by querying a large language model with a dedicated instruction (see Appendix~\ref{sec:appendix_rationale}). 

\subsection{Problem Generation Model}\label{sec:training}
Through concept extraction and rationale generation, we construct a dataset $\mathcal{D}=\{(\mathbf{c}, z, x)\}$ from the seed prompts, where $x$ represents a problem, $\mathbf{c}$ denotes the foundational concepts relevant to $x$, and $z$ captures the underlying thought process behind $x$. An LLM is then fine-tuned on $\mathcal{D}$ as a problem generation model, enabling it to jointly synthesize a rationale and the corresponding problem given a set of foundational concepts. The training objective is defined by Maximum Likelihood Estimation (MLE). 

To further ensure that synthesized problems are of high quality, we implement an iterative rejection sampling mechanism using the fine-tuned problem generation model from the previous step. Specifically, for each concept set $\mathbf{c} \in \mathcal{D}$, the model generates candidate rationale-problem pairs $(\tilde{z}, \tilde{x})$. These candidates undergo rigorous quality assessment through two independent LLMs serving as evaluators, which assign ratings based on predetermined criteria (detailed evaluator instructions are provided in Appendix~\ref{sec:appendix_evaluator}). Only candidate pairs receiving unanimous ``perfect'' ratings from both evaluators are retained for subsequent model optimization, thereby ensuring that the fine-tuning process exclusively incorporates exemplars of the highest quality. Algorithm~\ref{alg:prompt_generation} presents a formal summary of the rationale-guided problem generation procedure.

At inference time, the fine-tuned problem generation model is supplied with an arbitrary set of foundational concepts sampled uniformly from the entire training corpus. It then generates rationale-problem pairs $(z',x')$ that adhere to the rigorous quality standards established during training. This approach enables the generation of a large corpus of high-quality and challenging math problems that are robust and well-grounded in the underlying concepts. 

When using the synthesized problems for LLM reasoning, we leverage a powerful teacher model (details are presented in Section \ref{sec:exp}) to generate detailed, step-by-step solutions for each problem, thereby constructing a comprehensive training dataset $\mathcal{D}_{\text{train}} = \{(x', s')\}$, where $s'$ denotes the solution corresponding to problem $x'$, for training downstream mathematical reasoning models.

%% file: 3-experiments.tex
\begin{table*}[!ht]
\centering
\resizebox{0.8\textwidth}{!}{
\begin{tabular}{lcccc}
\toprule
\textbf{Models} & \textbf{Base} & \textbf{GSM8K} & \textbf{MATH-500} & \textbf{AIME2024} \\ 
\midrule
\multicolumn{5}{c}{\textit{short-CoT}} \\
Qwen2.5-Math-1.5B-Instruct & - & 84.8 & 75.8 & 10.0 \\ 
Openmathinstruct-1.5B & Qwen2.5-Math-1.5B & 84.9 & 63.8 & 13.3 \\ 
NuminaMath-1.5B &Qwen2.5-Math-1.5B & 85.1 & 69.0 & 16.7 \\ 
Evol-Instruct-1.5B &Qwen2.5-Math-1.5B & 84.6 & 67.4 & 10.0 \\ 
KPDDS-1.5B &Qwen2.5-Math-1.5B & 83.4 & 64.4 & 6.7 \\ 
\rowcolor{gray!20} \ourmodel-Qwen-1.5B &Qwen2.5-Math-1.5B & \textbf{87.1} & \textbf{78.4} & \textbf{26.7} \\ 
\midrule
\multicolumn{5}{c}{\textit{long-CoT}} \\
DeepSeek-R1-Distill-Qwen-1.5B$^{\dagger}$ &- & 85.1 & 80.2 & 23.3 \\ 
\rowcolor{gray!20} \ourmodel-DS-1.5B &DeepSeek-R1-Distill-Qwen-1.5B & \textbf{86.1} & \textbf{85.0} & \textbf{43.3} \\ 
\bottomrule
\end{tabular}
}
% \vspace{-2mm}
\caption{Evaluation results across three mathematical reasoning benchmarks for models with 1.5B parameters. Bold numbers indicate the highest performance in each respective setting. $\dagger$ indicates results reproduced using our prompt.}
% \vspace{-2mm}
\label{tab:main_results_1}
\end{table*}

\begin{table*}[!ht]
\centering
\resizebox{0.8\textwidth}{!}{
\begin{tabular}{lcccc}
\toprule
\textbf{Models} & \textbf{Base} & \textbf{GSM8K} & \textbf{MATH-500} & \textbf{AIME2024} \\ 
\midrule
\multicolumn{5}{c}{\textit{short-CoT}} \\
Qwen2.5-Math-7B-Instruct & - & \textbf{95.2} & 83.6 & 13.3 \\ 
Openmathinstruct-7B & Qwen2.5-Math-7B & 92.0 & 79.6 & 10.0 \\ 
NuminaMath-7B &Qwen2.5-Math-7B & 92.9 & 81.8 & 20.0 \\ 
Evol-Instruct-7B &Qwen2.5-Math-7B & 88.5 & 77.4 & 16.7 \\ 
KPDDS-7B &Qwen2.5-Math-7B & 89.9 & 76.0 & 10.0 \\ 
\rowcolor{gray!20} \ourmodel-Qwen-7B &Qwen2.5-Math-7B & 93.3 & \textbf{84.0} & \textbf{26.7} \\ 
\midrule
\multicolumn{5}{c}{\textit{long-CoT}} \\
DeepSeek-R1-Distill-Qwen-7B$^{\dagger}$ &- & 91.7 & 91.6 & 43.3 \\ 
\rowcolor{gray!20} \ourmodel-DS-7B &DeepSeek-R1-Distill-Qwen-7B & \textbf{92.6} & \textbf{93.0} & \textbf{60.0} \\ 
\bottomrule
\end{tabular}
}
% \vspace{-2mm}
\caption{Evaluation results across three mathematical reasoning benchmarks for models with 7B parameters. Bold numbers indicate the highest performance in each respective setting. $\dagger$ indicates results reproduced using our prompt.}
\label{tab:main_results_2}
% \vspace{-2mm}
\end{table*}

% \vspace{-2mm}
\section{Experiments}\label{sec:exp}

% \vspace{-2mm}
\subsection{Datasets and Evaluation Metrics}

We employ three standard benchmarks focused on mathematical reasoning. Specifically, we use the following datasets:
(1) \textbf{GSM8K}~\citep{cobbe2021training} is a dataset consisting of grade-school level math word problems that require logical reasoning. It tests a model’s ability to solve elementary-level math problems; (2) \textbf{MATH-500}~\citep{lightman2023let} is a dataset containing high school-level math problems. It serves to assess a model's ability to handle more advanced mathematical reasoning; and (3) \textbf{AIME 2024}~\citep{aimedata} is a benchmark that includes particularly challenging math problems from the American Invitational Mathematics Examination (AIME), designed to assess advanced problem-solving skills. These problems are significantly more difficult than typical high school-level math problems, requiring advanced reasoning and problem-solving strategies.
We use exact match accuracy as the primary metric for evaluating the performance of our method in math problem-solving tasks. Specifically, for these tasks, accuracy is determined by comparing the predicted final answer, enclosed by \texttt{\textbackslash boxed{}}, with the ground-truth answer.

\subsection{Baseline Methods}
We evaluate the proposed method in two settings: \textit{short-CoT} and \textit{long-CoT}. Short-CoT refers to the vanilla CoT~\citep{wei2022chain}, where reasoning tokens are generated before deriving the final answer. Long-CoT~\citep{openai-o1,guo2025deepseek}, on the other hand, requires the model to generate a deep reasoning process, which may include self-reflections, prior to producing the final CoT sequence.
For the short-CoT setting, we compare our method with the following problem generation baselines:
(1) \textbf{Evol-Instruct}: This method~\citep{luo2023wizardmath} aims to enhance the quality of instruction data by improving both its complexity and diversity, thus facilitating the generation of more varied and challenging problems;
(2) \textbf{KPDDS}: A data synthesis framework~\citep{huang2024key} that generates question-answer pairs by leveraging key concepts and exemplar practices derived from authentic data sources;
(3) \textbf{OpenMathInstruct}: This method~\citep{toshniwal2024openmathinstruct} utilizes few-shot learning to prompt an LLM to create new math problems based on existing examples, without explicit instructions for adjusting difficulty or introducing new constraints;
and (4) \textbf{NuminaMath}: This approach~\citep{li2024numinamath} uses an LLM to generate novel math questions starting from a reference problem.
Additionally, we compare with an open-source model \textbf{Qwen2.5-Math-Instruct}, which is known for its state-of-the-art performance  without relying on long-CoT reasoning.
For methods that have not released the generated problems, specifically Evol-Instruct and KPDDS, we follow their papers and use Llama-3.1-70B-Instruct~\citep{dubey2024llama} as the LLM to generate the problems, ensuring that the number of generated problems is consistent with ours. For NuminaMath~\footnote{\url{https://huggingface.co/AI-MO}} and OpenMathInstruct~\footnote{\url{https://huggingface.co/datasets/nvidia/OpenMathInstruct-2}}, we directly use the published problem sets. 
For all problem generation baselines, we use Qwen2.5-Math-72B-Instruct as the teacher model to generate solutions, ensuring a fair comparison across methods. 
For the long-CoT setting, we compare the proposed method with:
\textbf{DeepSeek-R1-Distill-Qwen}, a leading model~\citep{guo2025deepseek} that incorporates long-CoT reasoning~\citep{openai-o1}, allowing for more thorough and comprehensive problem-solving processes that generate solutions with deeper reasoning.

% \vspace{-1mm}
\subsection{Implementation Details}
We implement \ourmodel in two configurations: (1) \textbf{\ourmodel-Qwen}, distilled from Qwen2.5-Math-72B-Instruct~\citep{yang2024qwen2}, which is capable of generating short-CoT (vanilla Chain-of-Thought) reasoning, and (2) \textbf{\ourmodel-DS}, distilled from DeepSeek-R1-Distill-Qwen-7B~\citep{guo2025deepseek} (i.e., the teacher model), which is designed to generate long-CoT reasoning~\citep{openai-o1}. For \textbf{\ourmodel-Qwen}, we generate a total of $m=905,459$ problems, while for \textbf{\ourmodel-DS}, we generate $m=114,763$ problems\footnote{We generate significantly fewer problems in the long-CoT setting than in the shot-CoT setting, as solution generation in the long-CoT setting is considerably more costly.}.
For concept extraction, we construct a dataset consisting of $6,365$ seed prompts. We use Llama-3.1-70B-Instruct to extract the relevant concepts, with the number of concepts per problem set to $k=5$.
In the rationale generation phase, we leverage Llama-3.1-70B-Instruct~\citep{dubey2024llama}, Qwen2.5-72B-Instruct~\citep{yang2024qwen25llm}, and Qwen2.5-32B-Instruct to diversify the dataset, resulting in a total of $19,095$ concept-rationale-problem triples.
We initialize the problem generation model with Llama-3.1-8B. During the MLE training stage, we employ a learning rate of $2 \times 10^{-5}$ and a batch size of $64$. 
In the rejection sampling phase, we perform three rounds of evaluation, using Llama-3.1-70B-Instruct and Qwen2.5-72B-Instruct as evaluators. For the rejection sampling process, we continue to use the same learning rate of $2 \times 10^{-5}$ and batch size of $64$ as in the MLE training stage.
All experiments are conducted on 8×A100 80GB machines.

% \vspace{-1mm}
\subsection{Main Results}
The results of our experiments, presented in Tables~\ref{tab:main_results_1} and~\ref{tab:main_results_2}, reveal the following key insights: (1) Our method achieves state-of-the-art performance across multiple benchmarks, outperforming the baselines on both short-CoT and long-CoT settings. This highlights the efficacy of our rationale-driven approach in generating high-quality problems; (2) As the problem difficulty increases from GSM8K to MATH-500 and AIME2024, our method's advantage becomes more pronounced. This is particularly evident on AIME2024, which demands more advanced reasoning and problem construction. Our approach excels at generating Olympiad-level problems, capturing the more complex reasoning needed for such tasks; and (3) Our method further enhances the performance of long-CoT models like DeepSeek-R1-Distill-Qwen. The generation of Olympiad-level problems, with their higher complexity, taps into the full potential of long-CoT reasoning, enabling deeper and more effective reasoning processes.

\begin{table}[!ht]
\centering
\resizebox{1.0\columnwidth}{!}{
\begin{tabular}{lccc}
\toprule
\textbf{Models} & \textbf{GSM8K} & \textbf{MATH-500} & \textbf{AIME2024} \\ 
\midrule
 \ourmodel (full)  & \textbf{87.1} & \textbf{78.4} & \textbf{26.7} \\ 
\midrule
- rationale & 82.3 & 67.0 & 10.0 \\ 
- optimal  & 86.3 & 72.6 & 16.7 \\ 
- rejection sampling  & 85.9 & 75.2 & 20.0 \\ 
\bottomrule
\end{tabular}
}
% \vspace{-2mm}
\caption{Ablation study results for 1.5B parameter models. Bold numbers indicate the highest performance.}
% \vspace{-2mm}
\label{tab:ablation}
\end{table}

% \vspace{-2mm}
\section{Discussions}
In addition to the extensive evaluation across multiple benchmarks, we seek to further understand the underlying mechanisms of \ourmodel. Specifically, we explore the following research questions: (1) \textbf{RQ1:} How do the different components of \ourmodel contribute to its performance? (2) \textbf{RQ2:} How does the difficulty of the problems generated by \ourmodel compare to those from typical published works and open-source datasets? (3) \textbf{RQ3:} How does \ourmodel compare to state-of-the-art models with larger parameter sizes? (4) \textbf{RQ4:} What are the scaling properties of the problems generated by \ourmodel?

% \vspace{-1mm}
\subsection{Ablation Study for RQ1}\label{sec:ablation}
We perform an ablation study using 1.5B parameter models in the short-CoT setting and evaluate three variants of \ourmodel: exclusion of the rationale, denoted as ``- rationale''; exclusion of the optimality condition for the rationale, referred to as ``- optimal'';\footnote{Upon implementation, we removed the two conditions marked as ``(IMPORTANT)'' in the instruction for rationale generation~(Appendix~\ref{sec:appendix_rationale})} and exclusion of the rejection sampling process, represented as ``- rejection sampling''.

The results in Table~\ref{tab:ablation} show that the full version of \ourmodel consistently outperforms all variants, emphasizing the importance of each component. Excluding the rationale leads to the largest performance drop, indicating the crucial role of the rationale in guiding problem generation. The ``- optimal'' variant also experiences a performance decrease, albeit smaller, highlighting the significance of the rationale's optimal construction. The exclusion of rejection sampling has the least impact, suggesting that while it helps align the problem generation model with predefined quality criteria, the difficulty of problems plays a more important role in reasoning tasks.

\begin{figure*}[!ht]
    \centering
    \includegraphics[width=0.8\textwidth]{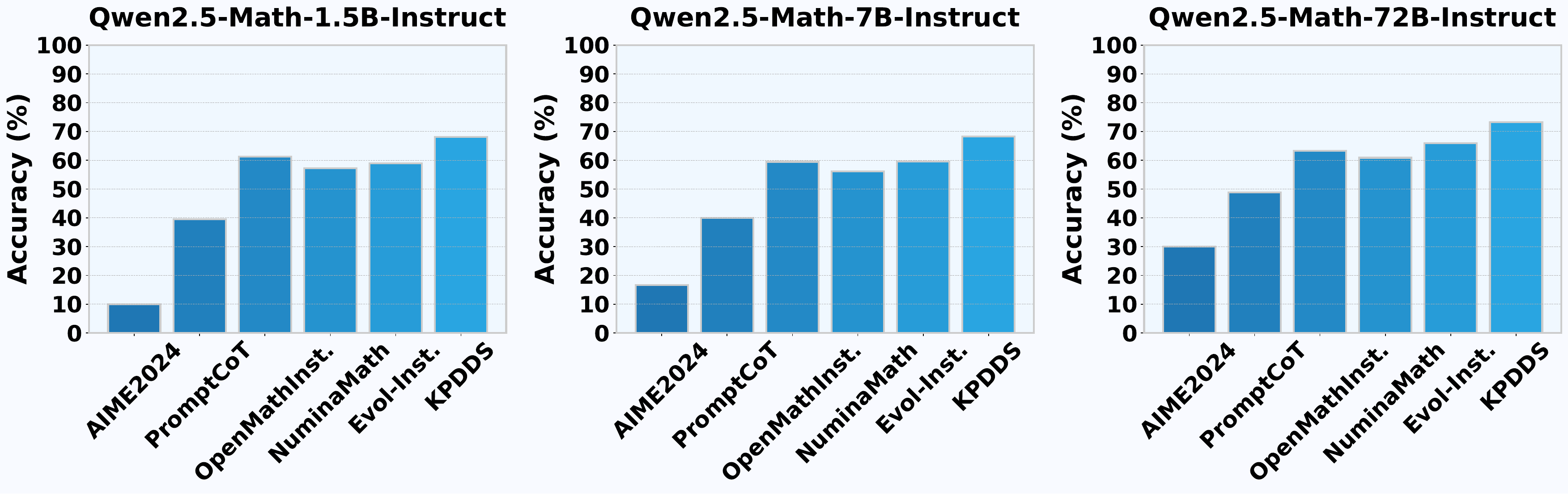} 
    % \vspace{-4mm}
    \caption{Performance comparison of models with varying capabilities, using Qwen2.5-Math series models, across problems from diverse sources.}
    % \vspace{-4mm}
    \label{fig:difficulty} 
\end{figure*}

% \vspace{-1mm}
\subsection{Analysis of Problem Difficulty for RQ2}
While Table \ref{tab:introduction} in Section \ref{sec:intro} has presented analysis on problem difficulty, we provide more details here. We assess the performance of Qwen2.5-Math-Instruct when tested on problem sets produced by different problem generation methods, including \ourmodel and other baselines. Specifically, we calculate the accuracy of Qwen2.5-Math-Instruct on these problem sets to compare their difficulty. We use the accuracy on AIME2024 as a reference. For open-source datasets, including NuminaMath and OpenMathInstruct, we use the provided answers in the datasets as the ground truth to calculate accuracy. For other methods, including KPDDS, Evol-Instruct, and \ourmodel, we use a strong reasoning model, DeepSeek-R1-Distill-Qwen-7B, to label the ground truth answers, applying self-consistency~\citep{wang2022self} with 8 rollouts to ensure answer validity. The results in Figure~\ref{fig:difficulty} indicate that the difficulty of the problems generated by \ourmodel is closer to AIME, outperforming the other methods by a significant margin.

Additionally, we compare the reasoning tokens required for different problems. To compute the number of reasoning tokens, we use the DeepSeek-R1-Distill-Qwen-7B model.
In this experiment, reasoning tokens include both the portion between ``<think>'' and ``</think>'' tags, as well as the final CoT. The results in Table~\ref{tab:introduction} demonstrate that \ourmodel-generated problems require a considerably higher number of reasoning tokens compared to the other methods, reflecting the increased difficulty of the problems generated by \ourmodel.

\begin{table}[!ht]
\centering
\resizebox{1.0\columnwidth}{!}{
\begin{tabular}{lcc}
\toprule
\textbf{Models} & \textbf{MATH-500} & \textbf{AIME2024} \\ 
\midrule
QwQ-32B &  90.6 & 50.0 \\ 
DeepSeek-R1-Distill-Qwen-32B$^{\dagger}$ & 90.4 & 63.3 \\
S1-32B & 93.0 & 56.7 \\ \midrule
\ourmodel-DS-7B  & 93.0 & 60.0  \\ 
\bottomrule
\end{tabular}
}
% \vspace{-2mm}
\caption{Performance comparison with state-of-the-art models having 32B parameters on MATH-500 and AIME2024. $\dagger$ indicates results reproduced using our prompt.}
% \vspace{-2mm}
\label{tab:32b}
\end{table}

% \vspace{-1mm}
\subsection{Performance Comparison for RQ3}
We further compare \ourmodel-DS-7B with state-of-the-art reasoning models that have significantly larger parameter sizes to evaluate whether our model can match the reasoning capabilities typically associated with larger models. Specifically, we compare \ourmodel-DS-7B with the models QwQ~\citep{qwq}, DeepSeek-R1-Distill-Qwen-32B~\citep{guo2025deepseek}, and S1-32B~\citep{muennighoff2025s1simpletesttimescaling}, all of which leverage long-CoT to solve problems. The results presented in Table~\ref{tab:32b} show that \ourmodel-DS-7B achieves performance comparable to the state-of-the-art 32B parameter models. This suggests that our approach, through synthesizing Olympiad-level problems, enables smaller models to perform at a level previously associated with much larger models, highlighting the effectiveness of our method in enhancing reasoning capabilities without requiring an increase in parameter size.

\begin{figure}[!ht]
\centering
\resizebox{0.8\columnwidth}{!}{\includegraphics{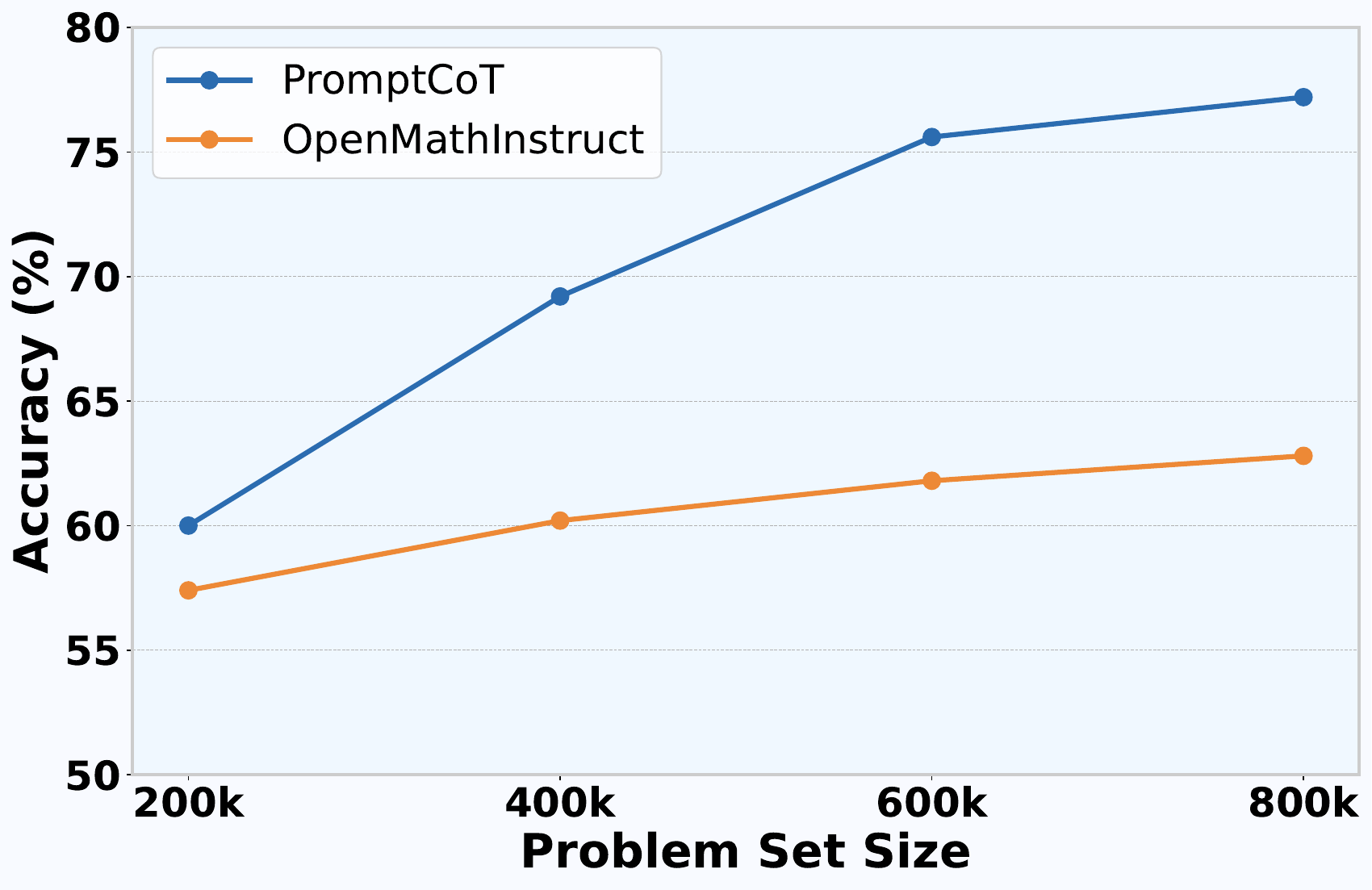}}
% \vspace{-4mm}
\caption{Comparison of Scaling Performance between \ourmodel and OpenMathInstruct across Varying Problem Set Sizes}
% \vspace{-4mm}
\label{fig:problem_size}
\end{figure}

% \vspace{-1mm}
\subsection{Scaling Properties for RQ4}

We evaluate the scaling property of \ourmodel by comparing it with OpenMathInstruct across varying problem sizes, ranging from 200K to 800K problems. To conduct the evaluation, we use Qwen2.5-Math-1.5B as the base model and Qwen2.5-Math-Instruct-72B to generate solutions for the problems. We select MATH-500 as the evaluation dataset due to its balanced difficulty and appropriate scale. 
The results shown in Figure~\ref{fig:problem_size} demonstrate that \ourmodel consistently outperforms OpenMathInstruct across all problem sizes. \ourmodel exhibits significant scalability, maintaining superior performance as the dataset size increases, while OpenMathInstruct's performance gradually plateaus with larger problem sets.

%% file: 4-related.tex
\section{Related Work}

\paragraph{Mathematical Reasoning with LLMs.}
Recent advances in LLMs' mathematical capabilities have been driven by three key directions: data enrichment, methodological innovation, and test-time scaling. While extensive mathematical pre-training corpora~\cite{paster2023openwebmathopendatasethighquality,wang2024mathpilebilliontokenscalepretrainingcorpus,azerbayev2024llemmaopenlanguagemodel,shao2024deepseekmath} and curated fine-tuning datasets~\cite{yue2023mammothbuildingmathgeneralist,yue2024mammoth2scalinginstructionsweb,li2024common7blanguagemodels,toshniwal2024openmathinstruct,wen2025unlocking,ye2025limoreasoning} have enhanced model performance, concerns persist about their true reasoning capabilities versus pattern memorization~\cite{li2024gsm,mirzadeh2024gsmsymbolicunderstandinglimitationsmathematical,zhang2024carefulexaminationlargelanguage,Kambhampati_2024}. To address these limitations, researchers have explored in-context learning~\citep{zhao2024subgoal}, advanced prompting techniques~\citep{cot,press2023measuring,imani2023mathprompter,zhao2024bba}, specialized training approaches~\citep{zhao2024sego,mitra2024orca,openai2024o1preview}, and tool integration~\citep{gao2023pal,schick2024toolformer}. Notable breakthroughs include DeepSeek-R1's~\cite{guo2025deepseek} rule-based reinforcement learning and test-time scaling advances~\cite{wu2024inference,guan2025rstar,muennighoff2025s1simpletesttimescaling}. However, generating Olympiad-level problems remains challenging.

\paragraph{Prompt Synthesis.}
Current prompt synthesis approaches fall into three categories: LLM-driven generation, data-derived synthesis, and heuristic composition. LLM-driven methods use advanced models to generate prompts from seeds~\cite{wang2023selfinstructaligninglanguagemodels,li2024numinamath}, sometimes incorporating personas~\cite{ge2024scalingsyntheticdatacreation} or concepts~\cite{huang2024key,tang2024mathscale}. \citet{xu2024magpiealignmentdatasynthesis} introduced auto-regressive template completion for query generation. Data-derived approaches extract prompts from existing sources, using techniques like instruction backtranslation~\cite{li2024selfalignmentinstructionbacktranslation,zhao2024subgoalxl} and web corpus refinement~\cite{yue2024mammoth2scalinginstructionsweb}. Heuristic composition methods systematically combine simpler elements, as demonstrated in visual reasoning~\cite{cheng2024mostbuildingplugandplayvisual} and mathematical problem construction~\cite{luo2023wizardmath,yu2024metamathbootstrapmathematicalquestions}. While these methods show promise, they lack comprehensive rationales in synthesis, limiting their ability to generate sophisticated problems. Our approach addresses this by explicitly incorporating rationales during synthesis.

%% file: 5-conclusion.tex
\section{Conclusion}
In this paper, we introduce \ourmodel, a novel approach for synthesizing Olympiad-level math problems. Our method integrates rationale generation to guide problem synthesis, inspired by the success of LLMs in reasoning tasks such as chain-of-thought. By leveraging this rationale and underlying mathematical concepts, we generate complex, high-quality problems that are well-suited to improve the reasoning capabilities of LLMs. Through extensive experimentation across multiple mathematical reasoning benchmarks, including GSM8K, MATH-500, and AIME2024, we show that \ourmodel outperforms existing problem generation methods.
% Additionally, our method demonstrates strong scalability, consistently maintaining performance as problem set sizes increase.

\section*{Ethical Considerations}
In accordance with the established Code of Ethics, this study utilizes only publicly available data and information, ensuring that no private or confidential resources are involved.

\section*{Limitations}

While \ourmodel makes significant strides in the generation of challenging mathematical problems, there are several limitations that should be addressed in future work to fully unlock its potential:

(1) The problem generation model used in \ourmodel is based on Llama-3.1-8B, which, due to its relatively smaller scale, may still struggle to generate exceptionally challenging problems, such as those seen in the International Mathematical Olympiad (IMO). Although it performs well for tasks within the scope of current benchmarks, its capacity limits the generation of problems requiring more advanced reasoning and complexity. In future work, we plan to explore the use of larger pre-trained models to improve the quality and difficulty of the generated problems, enabling the synthesis of more sophisticated problem sets.

(2) While \ourmodel has successfully scaled problem generation to 905K problems, this scale remains modest compared to the vast amounts of data used for LLM pretraining. Expanding the problem generation capacity to larger scales is crucial for generating more diverse and challenging problem sets.  Future research should focus on extending the scalability of problem generation to better align with the scale of current LLM training data, contributing to the development of more robust and capable models.

%% file: 6-appendix.tex
\appendix
\onecolumn

\section{Instruction for Concept Extraction}
\label{sec:appendix_concept}

The following prompt extracts domain-specific concepts from each seed prompt. This instruction directs the large language model to identify the salient concepts underlying the given mathematical problem.

\begin{tcolorbox}[colback=blue!5!white, colframe=blue!75!black, title=Concept Extraction Prompt]
As an expert in educational assessment, analyze this problem:

\{problem\}

Break down and identify \{num\_concepts\} foundational concepts being tested. List these knowledge points that:
\begin{itemize}
    \item Are core curriculum concepts typically taught in standard courses,
    \item Are precise and measurable (not vague like "understanding math"),
    \item Are essential building blocks needed to solve this problem,
    \item Represent fundamental principles rather than problem-specific techniques.
\end{itemize}

Think through your analysis step by step, then format your response as a Python code snippet containing a list of \{num\_concepts\} strings, where each string clearly describes one fundamental knowledge point.
\end{tcolorbox}

\section{Instruction for Rationale Generation}
\label{sec:appendix_rationale}

The following prompt is used to guide the large language model in generating a detailed thought process that serves as the rationale for prompt generation. Note that we emphasize two requirements with ``\textbf{(IMPORTANT)}'' markers in the prompt, aiming to enhance $p(x|z, \mathbf{c})$ and $p(z | \mathbf{c})$, respectively.

\begin{tcolorbox}[colback=blue!5!white, colframe=blue!75!black, title=Rationale Generation Instruction]
Imagine you are an expert in educational problem design.

You will be shown these components:

\textbf{Problem:} \{problem\}

\textbf{Fundamental Concepts:} \{list of concepts\}

\textbf{Difficulty Level:} \{difficulty\_level\}

Your task is to reverse-engineer a clear thinking process that shows how a teacher might design this problem. This thinking process should:
\begin{itemize}
    \item Show how combining the given foundational concepts naturally leads to a problem at the specified difficulty level.
    \item Include all key decisions and reasoning that shaped the problem design.
    \item \textbf{(IMPORTANT)} Be so precise and detailed that another teacher following these exact steps would recreate the identical problem.
    \item \textbf{(IMPORTANT)} Be so natural and logical that another teacher could derive the same thinking process using only the foundational concepts and difficulty level.
\end{itemize}

Present your answer after ``Thinking Process:'' with the complete step-by-step thinking process described above.
\end{tcolorbox}

\section{Instruction for Rejection Sampling}\label{sec:appendix_evaluator}

\begin{tcolorbox}[colback=blue!5!white, colframe=blue!75!black, title=Evaluation Prompt]
\small
As a critical expert in educational problem design, evaluate the following problem components:

\textbf{=== GIVEN MATERIALS ===}

\begin{itemize}
    \item \textbf{1. Problem \& Design Rationale:} \texttt{\{rationale\_and\_problem\}} \\
    (The rationale describes the author's thinking process and justification in designing this problem)
    \item \textbf{2. Foundational Concepts:} \texttt{\{concept\_text\}}
    \item \textbf{3. Target Difficulty Level:} \texttt{\{level\}}
\end{itemize}

\textbf{=== EVALUATION CRITERIA ===}

Rate each criterion as: [Perfect | Acceptable | Bad]

\begin{itemize}
    \item \textbf{1. FORMAT}
    \begin{itemize}
        \item Verify correct implementation of markup tags:
        \item \texttt{<!-- BEGIN RATIONALE --> [design thinking process] <!-- END RATIONALE -->}
        \item \texttt{<!-- BEGIN PROBLEM --> [problem] <!-- END PROBLEM -->}
    \end{itemize}

    \item \textbf{2. FACTUAL ACCURACY}
    \begin{itemize}
        \item Check for any incorrect or misleading information in both problem and rationale
        \item Verify mathematical, scientific, or logical consistency
    \end{itemize}

    \item \textbf{3. DIFFICULTY ALIGNMENT}
    \begin{itemize}
        \item Assess if problem complexity matches the specified difficulty level
        \item Evaluate if cognitive demands align with target level
    \end{itemize}

    \item \textbf{4. CONCEPT COVERAGE}
    \begin{itemize}
        \item Evaluate how well the problem incorporates the given foundational concepts
        \item Check for missing concept applications
    \end{itemize}

    \item \textbf{5. SOLVABILITY}
    \begin{itemize}
        \item Verify if the problem has at least one valid solution
        \item Check if all necessary information for solving is provided
    \end{itemize}
\end{itemize}

\textbf{=== RESPONSE FORMAT ===}

For each criterion, provide:
\begin{itemize}
    \item Rating: [Perfect | Acceptable | Bad]
    \item Justification: Clear explanation for the rating
\end{itemize}

\textbf{=== FINAL VERDICT ===}

After providing all criterion evaluations, conclude your response with:
\begin{quote}
'Final Judgement: [verdict]'
\end{quote}
where verdict must be one of:
\begin{itemize}
    \item 'perfect' (if both FACTUAL ACCURACY and SOLVABILITY are Perfect, at least two other criteria are Perfect, and no Bad ratings)
    \item 'acceptable' (if no Bad ratings and doesn't qualify for perfect)
    \item 'bad' (if ANY Bad ratings)
\end{itemize}

Note: The 'Final Judgement: [verdict]' line must be the final line of your response.
\end{tcolorbox}

\section{Proof of the Optimal Variational Distribution}
\label{sec:appendix_proof}

In this section, we provide a rigorous derivation of the optimal variational distribution \(q^\star(z \mid \mathbf{c}, x)\) that maximizes the evidence lower bound (ELBO)
\[
\mathcal{L}(q) = \mathbb{E}_{q(z \mid \mathbf{c}, x)} \left[ \log \frac{p(x, z \mid \mathbf{c})}{q(z \mid \mathbf{c}, x)} \right]
\]
subject to the normalization constraint
\[
\sum_{z} q(z \mid \mathbf{c}, x) = 1.
\]

We wish to maximize the functional
\[
\mathcal{L}(q) = \sum_{z} q(z \mid \mathbf{c}, x) \log \frac{p(x, z \mid \mathbf{c})}{q(z \mid \mathbf{c}, x)},
\]
with respect to \(q(z \mid \mathbf{c}, x)\), subject to
\[
\sum_{z} q(z \mid \mathbf{c}, x) = 1.
\]
To incorporate the constraint, we define the Lagrangian functional
\[
\mathcal{J}(q, \lambda) = \sum_{z} q(z \mid \mathbf{c}, x) \log \frac{p(x, z \mid \mathbf{c})}{q(z \mid \mathbf{c}, x)} + \lambda \left( \sum_{z} q(z \mid \mathbf{c}, x) - 1 \right),
\]
where \(\lambda\) is a Lagrange multiplier.

For each \(z\), we take the derivative of \(\mathcal{J}(q, \lambda)\) with respect to \(q(z \mid \mathbf{c}, x)\). Using standard calculus of variations, we obtain:
\[
\frac{\partial \mathcal{J}}{\partial q(z \mid \mathbf{c}, x)} = \log \frac{p(x, z \mid \mathbf{c})}{q(z \mid \mathbf{c}, x)} - 1 + \lambda.
\]
Setting this derivative to zero for optimality, we have
\[
\log \frac{p(x, z \mid \mathbf{c})}{q(z \mid \mathbf{c}, x)} - 1 + \lambda = 0.
\]

Rearrange the above equation to isolate \(q(z \mid \mathbf{c}, x)\):
\[
\log \frac{p(x, z \mid \mathbf{c})}{q(z \mid \mathbf{c}, x)} = 1 - \lambda.
\]
Exponentiating both sides yields
\[
\frac{p(x, z \mid \mathbf{c})}{q(z \mid \mathbf{c}, x)} = e^{1 - \lambda},
\]
or equivalently,
\[
q(z \mid \mathbf{c}, x) = p(x, z \mid \mathbf{c}) \, e^{-(1 - \lambda)}.
\]

We now enforce the normalization constraint:
\[
\sum_{z} q(z \mid \mathbf{c}, x) = e^{-(1 - \lambda)} \sum_{z} p(x, z \mid \mathbf{c}) = e^{-(1 - \lambda)} \, p(x \mid \mathbf{c}) = 1.
\]
Solving for \(e^{-(1 - \lambda)}\), we obtain
\[
e^{-(1 - \lambda)} = \frac{1}{p(x \mid \mathbf{c})}.
\]

Substituting this result back into the expression for \(q(z \mid \mathbf{c}, x)\), we arrive at the optimal variational distribution:
\[
q^\star(z \mid \mathbf{c}, x) = \frac{p(x, z \mid \mathbf{c})}{p(x \mid \mathbf{c})}.
\]
This completes the proof.

\section{Case Study}
To further demonstrate the effectiveness of \ourmodel, we visualize the step-by-step rationale generation and the corresponding problem produced by our problem generation model~(as detailed in \S\ref{sec:training}) in Figure \ref{fig:case_study_rationale1} and \ref{fig:case_study_rationale2}, highlighting how the rationale is grounded in the problem's core elements and guides the model towards generating Olympiad-level problems.

Figures \ref{fig:case_study_7b_1} - \ref{fig:case_study_7b_5} show the solutions generated by \ourmodel-Qwen-7B and \ourmodel-DS-7B for various problem sources.

\begin{figure}[htb]
\small
    \centering
    \fbox{%
        \begin{minipage}{0.95\textwidth}
            \vspace{0.5em}
            \textbf{Foundational Concepts:}
            \begin{enumerate}
                \item Geometric arrangements and intersection points, including the concept of interior points created by intersecting lines
                \item Understanding of expected value in probability theory
                \item Understanding of ratios and proportions, particularly in the context of comparing areas of geometric shapes
                \item Ability to apply algebraic manipulations, such as solving linear equations and simplifying expressions, to solve problems
                \item Knowledge of integer arithmetic and the properties of integers, including powers of 2
            \end{enumerate}
            
            \vspace{1em}
            \textbf{Rationale:}
            
            \noindent\textbf{Step 1: Concept Selection and Combination} \\
            To design a problem at the AMC12 difficulty level, we need to combine multiple foundational concepts in a way that creates a challenging yet solvable problem. We start by selecting concepts that naturally fit together. In this case, we choose:
            \begin{itemize}
                \item Geometric arrangements and intersection points (Concept 1)
                \item Expected value in probability theory (Concept 2)
                \item Ratios and proportions (Concept 3)
                \item Algebraic manipulations (Concept 4)
                \item Integer arithmetic and properties of integers (Concept 5)
            \end{itemize}
            We want to create a problem that requires the student to apply these concepts in a cohesive and logical manner.
            
            \vspace{0.5em}
            \noindent\textbf{Step 2: Problem Theme and Context} \\
            To make the problem more engaging, we decide to use a geometric theme involving intersecting lines. We choose a square as the geometric shape, as it is simple and familiar to students. We also introduce the idea of interior points created by intersecting lines, which will be used to calculate expected values.
            
            \vspace{0.5em}
            \noindent\textbf{Step 3: Problem Statement and Parameters} \\
            We define the problem statement: "For each interior point $P$ of a square, let $a(P)$ denote the number of distinct intersection points of the lines that pass through $P$ and the sides of the square." This statement introduces the concept of intersection points and sets the stage for the problem.
            
            To add complexity, we introduce two specific points, $A$ and $B$, and ask for the ratio $\frac{a(A)}{a(B)}$. This requires students to apply their understanding of ratios and proportions (Concept 3).
            
            \vspace{0.5em}
            \noindent\textbf{Step 4: Expected Value and Probability} \\
            To incorporate expected value (Concept 2), we introduce the idea of a randomly chosen point $P$. We want students to think about the expected number of intersection points for a randomly chosen point, which will be used to calculate the ratio $\frac{a(A)}{a(B)}$.
            
            \vspace{0.5em}
            \noindent\textbf{Step 5: Algebraic Manipulations and Integer Arithmetic} \\
            To make the problem more challenging, we decide to ask for the sum of the digits of the ratio $\frac{a(A)}{a(B)}$. This requires students to apply algebraic manipulations (Concept 4) and integer arithmetic (Concept 5) to simplify the expression and find the sum of its digits.
            
            \vspace{0.5em}
            \noindent\textbf{Step 6: Difficulty Level and Constraints} \\
            To ensure the problem is at the AMC12 difficulty level, we need to balance the complexity of the concepts and the calculations required. We decide to keep the problem statement concise and focused on the key concepts, rather than introducing additional complexity.
            
            \vspace{0.5em}
            \noindent\textbf{Step 7: Finalizing the Problem} \\
            After considering the above steps, we finalize the problem:
            
            "For each interior point $P$ of a square, let $a(P)$ denote the number of distinct intersection points of the lines that pass through $P$ and the sides of the square. For example, $a(P)$ has the values shown below for some points $P$ in the square. What is the sum of the digits of $\frac{a(A)}{a(B)}$?"
            
            This problem requires students to apply a range of concepts in a logical and cohesive manner, making it suitable for the AMC12 difficulty level.
            
            \vspace{1em}
            \textbf{Final Problem:} \\
            For each interior point $P$ of a square, let $a(P)$ denote the number of distinct intersection points of the lines that pass through $P$ and the sides of the square. For example, $a(P)$ has the values shown below for some points $P$ in the square. [Diagram showing a square with points $A$, $B$, $C$, $D$ and their corresponding $a(P)$ values] What is the sum of the digits of $\frac{a(A)}{a(B)}$?
        \end{minipage}%
    }
    \caption{Case study of rationale and problem generation.}
    \label{fig:case_study_rationale1}
\end{figure}

\begin{figure}[htb]
\small
    \centering
    \fbox{%
        \begin{minipage}{0.95\textwidth}
            \vspace{0.5em}
            \textbf{Foundational Concepts:}
            \begin{enumerate}
                \item Skill in solving inequalities involving integers
                \item Ability to recognize and analyze patterns in number sequences, including multiples of a number
                \item Understanding of the unit circle and the periodicity of trigonometric functions, especially within the interval from 0 to $2\pi$
                \item Understanding of the concept of collinearity, including the ability to determine whether three points lie on the same line in 3D space
                \item Ability to set up and solve algebraic equations to represent and solve problems
            \end{enumerate}
            
            \vspace{1em}
            \textbf{Rationale:}
            
            \noindent\textbf{Step 1: Selecting Foundational Concepts and Difficulty Level} \\
            I start by selecting the foundational concepts and the desired difficulty level for the problem. In this case, I choose the concepts of solving inequalities involving integers, recognizing patterns in number sequences, understanding the unit circle and periodicity of trigonometric functions, understanding collinearity, and setting up and solving algebraic equations. The desired difficulty level is HMMT-Nov, which suggests a challenging problem that requires careful analysis and application of multiple concepts.
            
            \vspace{0.5em}
            \noindent\textbf{Step 2: Brainstorming Ideas} \\
            Next, I brainstorm ideas that combine the selected concepts. I consider problems involving sequences of integers, trigonometric functions, and 3D geometry. I think about how these concepts can be connected, and I start to form ideas about problems that could require the application of multiple concepts.
            
            \vspace{0.5em}
            \noindent\textbf{Step 3: Focusing on Collinearity and Trigonometric Functions} \\
            I decide to focus on the concept of collinearity in 3D space, as it is a rich area for problem design. I think about how I can use trigonometric functions to create a scenario where three points are collinear. I consider the unit circle and the periodicity of trigonometric functions, and I realize that I can use the fact that the cosine and sine functions are periodic with a period of $2\pi$ to create a repeating pattern.
            
            \vspace{0.5em}
            \noindent\textbf{Step 4: Introducing a Sequence of Integers} \\
            To incorporate the concept of sequences of integers, I decide to introduce a sequence $a_n$ that is defined in terms of $n$. I consider how I can use the fact that $a_n$ is an integer to create a problem that requires careful analysis of the sequence. I think about how I can use the sequence to create a scenario where the three points are collinear.
            
            \vspace{0.5em}
            \noindent\textbf{Step 5: Creating a Scenario with Collinear Points} \\
            I decide to create a scenario where three points are collinear in 3D space, and the coordinates of the points are defined in terms of the sequence $a_n$. I think about how I can use the fact that the cosine and sine functions are periodic with a period of $2\pi$ to create a repeating pattern, and I realize that I can use this to create a scenario where the three points are collinear. I decide to use the coordinates $(x_n, y_n, z_n)$, where $x_n = \cos(a_n)$, $y_n = \sin(a_n)$, and $z_n = n$.
            
            \vspace{0.5em}
            \noindent\textbf{Step 6: Defining the Problem} \\
            I define the problem as finding the number of integers $1 \leq k \leq 2017$ such that the points $(x_k, y_k, z_k)$, $(x_{k+1}, y_{k+1}, z_{k+1})$, and $(x_{k+2}, y_{k+2}, z_{k+2})$ all lie on the same line. I realize that this problem requires the application of multiple concepts, including solving inequalities involving integers, recognizing patterns in number sequences, understanding the unit circle and periodicity of trigonometric functions, understanding collinearity, and setting up and solving algebraic equations.
            
            \vspace{0.5em}
            \noindent\textbf{Step 7: Finalizing the Problem} \\
            I finalize the problem by specifying the interval $0 \leq a_n < 2\pi$ and the condition that $a_n$ is an integer. I realize that this condition ensures that the problem is well-defined and that the solution can be found using careful analysis of the sequence $a_n$. I am confident that the problem is at the desired difficulty level, and I am excited to see how students will approach the problem.
            
            \vspace{1em}
            \textbf{Final Problem:} \\
            Let $a_{1}, a_{2}, \ldots$ be a sequence of integers such that $0 \leq a_{n}<2 \pi$ for all $n$, and such that there exists a positive integer $M$ for which $a_{n}=a_{n+M}$ for all $n$. For how many integers $1 \leq k \leq 2017$ do there exist real numbers $x_{k}, y_{k}, z_{k}$ such that the points $\left(x_{k}, y_{k}, z_{k}\right),\left(x_{k+1}, y_{k+1}, z_{k+1}\right)$, and $\left(x_{k+2}, y_{k+2}, z_{k+2}\right)$ all lie on the same line, and such that $x_{k}=\cos \left(a_{k}\right), y_{k}=\sin \left(a_{k}\right)$, and $z_{k}=k$ ?
        \end{minipage}%
    }
    \caption{Case study of rationale and problem generation.}
    \label{fig:case_study_rationale2}
\end{figure}

\begin{figure}[htb]
\small
    \centering
    \fbox{%
        \begin{minipage}{0.95\textwidth}
            \vspace{0.5em}
            \textbf{Problem:} \\
            Janet's ducks lay 16 eggs per day. She eats three for breakfast every morning and bakes muffins for her friends every day with four. She sells the remainder at the farmers' market daily for \$2 per fresh duck egg. How much in dollars does she make every day at the farmers' market?
            
            \vspace{1em}
            \textbf{Thought Process:} \\
            \noindent\texttt{<think>} \\
            Okay, so I need to figure out how much Janet makes every day from selling her duck eggs at the farmers' market. Let me try to break this down step by step.
            
            First, the problem says that Janet's ducks lay 16 eggs per day. That seems straightforward. So, she has 16 eggs each day that she can potentially sell or use for something else.
            
            Next, she eats three eggs for breakfast every morning. Hmm, so every day, she consumes 3 eggs. That means those 3 eggs aren't available for sale anymore. So, I should subtract those 3 eggs from the total number of eggs she lays each day. Let me write that down:
            
            Total eggs laid: 16 \\
            Eggs eaten for breakfast: 3 \\
            So, eggs left after breakfast: $16 - 3 = 13$
            
            Wait, hold on. Is that all? Because she also bakes muffins for her friends every day using four eggs. So, she uses another 4 eggs for baking muffins. That means I need to subtract those 4 eggs as well. Let me adjust my calculation:
            
            Eggs used for muffins: 4 \\
            So, total eggs used: $3$ (eaten) $+ 4$ (baked) $= 7$
            
            Therefore, the remaining eggs she can sell would be the total eggs minus the ones she ate and minus the ones she used for muffins. Let me compute that:
            
            Eggs sold = Total eggs - Eggs eaten - Eggs used for muffins \\
            Eggs sold = $16 - 3 - 4 = 9$
            
            So, she has 9 eggs left to sell each day. Now, she sells each fresh duck egg for \$2 at the farmers' market. To find out how much money she makes daily, I need to multiply the number of eggs she sells by the price per egg.
            
            Money made = Eggs sold $\times$ Price per egg \\
            Money made = $9 \times \$2 = \$18$
            
            Let me just double-check my calculations to make sure I didn't miss anything. She starts with 16 eggs. She eats 3, which leaves her with 13. Then she uses 4 for muffins, leaving her with 9. She sells each of those 9 for \$2, so $9 \times 2$ is indeed 18.
            
            Is there any other detail I might have overlooked? The problem mentions she lays 16 eggs per day, eats 3 for breakfast, bakes muffins with 4, and sells the rest. There doesn't seem to be any other usage or expense mentioned. So, it looks like all the eggs are either consumed or sold, no leftover eggs or additional costs.
            
            Therefore, I think my calculation is correct. She makes \$18 each day from selling her duck eggs at the farmers' market.
            
            \textbf{Final Answer} \\
            Janet makes $\boxed{18}$ dollars every day at the farmers' market. \\
            \texttt{</think>}
            
            \vspace{1em}
            \textbf{Solution:} \\
            Janet's ducks lay 16 eggs per day. She eats 3 eggs for breakfast every morning and uses 4 eggs to bake muffins daily. The remaining eggs are sold at the farmers' market for \$2 per egg.
            
            \begin{enumerate}
                \item Total eggs laid: 16
                \item Eggs eaten for breakfast: 3
                \item Eggs used for muffins: 4
                \item Eggs sold: $16 - 3 - 4 = 9$
                \item Money made from selling eggs: $9 \times 2 = 18$
            \end{enumerate}
            
            Thus, Janet makes $\boxed{18}$ dollars every day at the farmers' market.
        \end{minipage}%
    }
    \caption{Solution generated by \ourmodel-DS-7B for a GSM8K problem.}
    \label{fig:case_study_7b_1}
\end{figure}

\begin{figure}[htb]
\small
    \centering
    \fbox{%
        \begin{minipage}{0.95\textwidth}
            \vspace{0.5em}
            \textbf{Problem:} \\
            Jen enters a lottery by picking $4$ distinct numbers from $S=\{1,2,3,\cdots,9,10\}.$ $4$ numbers are randomly chosen from $S.$ She wins a prize if at least two of her numbers were $2$ of the randomly chosen numbers, and wins the grand prize if all four of her numbers were the randomly chosen numbers. The probability of her winning the grand prize given that she won a prize is $\tfrac{m}{n}$ where $m$ and $n$ are relatively prime positive integers. Find $m+n$.
            
            \vspace{1em}
            \textbf{Solution:}
            
            \noindent\textbf{1. Calculate the total number of ways to choose 4 numbers from $S$:} \\
            The set $S$ contains 10 elements, so the number of ways to choose 4 distinct numbers from $S$ is:
            \[
            \binom{10}{4} = \frac{10!}{4!6!} = 210
            \]
            
            \vspace{0.5em}
            \noindent\textbf{2. Calculate the number of ways Jen can win a prize:} \\
            Jen wins a prize if at least 2 of her numbers match the randomly chosen numbers. We will use the complement rule to find this. First, we calculate the number of ways she can win no prize or exactly 1 prize.
            
            \begin{itemize}
                \item \textbf{Number of ways she wins no prize:} \\
                She must choose 4 numbers from the 6 numbers that are not the randomly chosen numbers. The number of ways to do this is:
                \[
                \binom{6}{4} = \frac{6!}{4!2!} = 15
                \]
                
                \item \textbf{Number of ways she wins exactly 1 prize:} \\
                She must choose 1 number from the 4 randomly chosen numbers and 3 numbers from the 6 numbers that are not the randomly chosen numbers. The number of ways to do this is:
                \[
                \binom{4}{1} \binom{6}{3} = 4 \cdot \frac{6!}{3!3!} = 4 \cdot 20 = 80
                \]
            \end{itemize}
            
            Therefore, the number of ways she wins no prize or exactly 1 prize is:
            \[
            15 + 80 = 95
            \]
            So, the number of ways she wins a prize is:
            \[
            210 - 95 = 115
            \]
            
            \vspace{0.5em}
            \noindent\textbf{3. Calculate the number of ways Jen can win the grand prize:} \\
            Jen wins the grand prize if all 4 of her numbers match the randomly chosen numbers. The number of ways to do this is:
            \[
            \binom{4}{4} = 1
            \]
            
            \vspace{0.5em}
            \noindent\textbf{4. Calculate the probability of winning the grand prize given that she won a prize:} \\
            This is the ratio of the number of ways she can win the grand prize to the number of ways she can win a prize:
            \[
            \frac{1}{115}
            \]
            
            Since 1 and 115 are relatively prime, the probability is in its simplest form. Therefore, $m = 1$ and $n = 115$, and the sum $m + n$ is:
            \[
            1 + 115 = 116
            \]
            
            \vspace{0.5em}
            The final answer is:
            \[
            \boxed{116}
            \]
        \end{minipage}%
    }
    \caption{Solution generated by \ourmodel-Qwen-7B for an AIME2024 problem.}
    \label{fig:case_study_7b_2}
\end{figure}

\begin{figure}[htb]
\small
    \centering
    \fbox{%
        \begin{minipage}{0.95\textwidth}
            \vspace{0.5em}
            \textbf{Problem:} \\
            There exist real numbers $x$ and $y$, both greater than 1, such that $\log_x\left(y^x\right)=\log_y\left(x^{4y}\right)=10$. Find $xy$.
            
            \vspace{1em}
            \textbf{Solution:}
            
            To solve the problem, we start with the given equations:
            \[
            \log_x(y^x) = 10 \quad \text{and} \quad \log_y(x^{4y}) = 10
            \]
            
            \vspace{0.5em}
            \noindent\textbf{Step 1: Convert logarithmic equations to exponential form} \\
            For the first equation, $\log_x(y^x) = 10$, we have:
            \[
            x^{10} = y^x
            \]
            
            For the second equation, $\log_y(x^{4y}) = 10$, we have:
            \[
            y^{10} = x^{4y}
            \]
            
            \vspace{0.5em}
            \noindent\textbf{Step 2: Take natural logarithm of both sides} \\
            Starting with $x^{10} = y^x$, we get:
            \[
            10 \ln x = x \ln y
            \]
            Rearranging this equation, we obtain:
            \[
            \frac{\ln y}{\ln x} = \frac{10}{x}
            \]
            
            Taking the natural logarithm of both sides of $y^{10} = x^{4y}$, we get:
            \[
            10 \ln y = 4y \ln x
            \]
            Rearranging this equation, we obtain:
            \[
            \frac{\ln y}{\ln x} = \frac{4y}{10} = \frac{2y}{5}
            \]
            
            \vspace{0.5em}
            \noindent\textbf{Step 3: Set equal expressions} \\
            Since both expressions are equal to $\frac{\ln y}{\ln x}$, we can set them equal to each other:
            \[
            \frac{10}{x} = \frac{2y}{5}
            \]
            
            \vspace{0.5em}
            \noindent\textbf{Step 4: Solve for $xy$} \\
            Cross-multiplying gives:
            \[
            50 = 2xy
            \]
            Solving for $xy$, we get:
            \[
            xy = 25
            \]
            
            \vspace{0.5em}
            Thus, the value of $xy$ is $\boxed{25}$.
        \end{minipage}%
    }
    \caption{Solution generated by \ourmodel-Qwen-7B for an AIME2024 problem.}
    \label{fig:case_study_7b_3}
\end{figure}

\begin{figure}[htb]
\small
    \centering
    \fbox{%
        \begin{minipage}{0.95\textwidth}
            \vspace{0.5em}
            \textbf{Problem:} \\
            Alice chooses a set $A$ of positive integers. Then Bob lists all finite nonempty sets $B$ of positive integers with the property that the maximum element of $B$ belongs to $A$. Bob's list has 2024 sets. Find the sum of the elements of $A$.
            
            \vspace{1em}
            \textbf{Solution:}
            
            Let's denote the elements of $A$ as $a_1, a_2, \ldots, a_n$ where $a_1 < a_2 < \cdots < a_n$.
            
            \vspace{0.5em}
            \noindent\textbf{Step 1: Analyze the sets $B$} \\
            For each element $a_i$ in $A$, the sets $B$ that have $a_i$ as their maximum element are all subsets of $\{1, 2, \ldots, a_i\}$ that include $a_i$. The number of such subsets is $2^{a_i-1}$ because we can choose any subset of $\{1, 2, \ldots, a_i-1\}$ and add $a_i$ to it.
            
            \vspace{0.5em}
            \noindent\textbf{Step 2: Set up the equation} \\
            The total number of sets $B$ is the sum of $2^{a_i-1}$ for all $i$ from 1 to $n$:
            \[
            2^{a_1-1} + 2^{a_2-1} + \cdots + 2^{a_n-1} = 2024
            \]
            
            \vspace{0.5em}
            \noindent\textbf{Step 3: Convert 2024 to binary} \\
            \[
            2024_{10} = 11111101000_2
            \]
            This binary representation tells us that:
            \[
            2024 = 2^{10} + 2^9 + 2^8 + 2^7 + 2^6 + 2^5 + 2^3
            \]
            
            \vspace{0.5em}
            \noindent\textbf{Step 4: Identify the elements of $A$} \\
            From the binary representation, we can identify:
            \begin{align*}
                a_1-1 &= 3 \\
                a_2-1 &= 5 \\
                a_3-1 &= 6 \\
                a_4-1 &= 7 \\
                a_5-1 &= 8 \\
                a_6-1 &= 9 \\
                a_7-1 &= 10
            \end{align*}
            
            Therefore, the elements of $A$ are:
            \[
            a_1 = 4, a_2 = 6, a_3 = 7, a_4 = 8, a_5 = 9, a_6 = 10, a_7 = 11
            \]
            
            \vspace{0.5em}
            \noindent\textbf{Step 5: Calculate the sum} \\
            The sum of the elements of $A$ is:
            \[
            4 + 6 + 7 + 8 + 9 + 10 + 11 = 55
            \]
            
            \vspace{0.5em}
            Thus, the sum of the elements of $A$ is $\boxed{55}$.
        \end{minipage}%
    }
    \caption{Solution generated by \ourmodel-Qwen-7B for an AIME2024 problem.}
    \label{fig:case_study_7b_4}
\end{figure}

\begin{figure}[htb]
\small
    \centering
    \fbox{%
        \begin{minipage}{0.95\textwidth}
            \vspace{0.5em}
            \textbf{Problem:} \\
            Find the largest possible real part of $(75+117i)z + \frac{96+144i}{z}$ where $z$ is a complex number with $|z|=4$.
            
            \vspace{1em}
            \textbf{Solution:}
            
            \noindent\textbf{Step 1: Express $z$ in polar form} \\
            Let $z = 4e^{i\theta} = 4(\cos\theta + i\sin\theta)$ \\
            Then $\frac{1}{z} = \frac{1}{4e^{i\theta}} = \frac{1}{4}e^{-i\theta} = \frac{1}{4}(\cos\theta - i\sin\theta)$
            
            \vspace{0.5em}
            \noindent\textbf{Step 2: Expand first term} \\
            $(75+117i) \cdot 4(\cos\theta + i\sin\theta)$ \\
            $= 4(75\cos\theta + 75i\sin\theta + 117i\cos\theta + 117i^2\sin\theta)$ \\
            $= 4(75\cos\theta + 75i\sin\theta + 117i\cos\theta - 117\sin\theta)$ \\
            $= 4((75\cos\theta - 117\sin\theta) + (75\sin\theta + 117\cos\theta)i)$ \\
            $= (300\cos\theta - 468\sin\theta) + (300\sin\theta + 468\cos\theta)i$
            
            \vspace{0.5em}
            \noindent\textbf{Step 3: Expand second term} \\
            $(96+144i) \cdot \frac{1}{4}(\cos\theta - i\sin\theta)$ \\
            $= \frac{1}{4}(96\cos\theta - 96i\sin\theta + 144i\cos\theta - 144i^2\sin\theta)$ \\
            $= \frac{1}{4}(96\cos\theta - 96i\sin\theta + 144i\cos\theta + 144\sin\theta)$ \\
            $= \frac{1}{4}((96\cos\theta + 144\sin\theta) + (-96\sin\theta + 144\cos\theta)i)$ \\
            $= (24\cos\theta + 36\sin\theta) + (-24\sin\theta + 36\cos\theta)i$
            
            \vspace{0.5em}
            \noindent\textbf{Step 4: Add terms} \\
            $(300\cos\theta - 468\sin\theta + 24\cos\theta + 36\sin\theta)$ \\
            $+ (300\sin\theta + 468\cos\theta - 24\sin\theta + 36\cos\theta)i$ \\
            $= (324\cos\theta - 432\sin\theta) + (324\sin\theta + 504\cos\theta)i$
            
            \vspace{0.5em}
            \noindent\textbf{Step 5: Find maximum real part} \\
            The real part is $324\cos\theta - 432\sin\theta$ \\
            Maximum value of $a\cos\theta + b\sin\theta$ is $\sqrt{a^2 + b^2}$ \\
            Here, $a = 324$ and $b = -432$ \\
            Maximum value $= \sqrt{324^2 + (-432)^2}$ \\
            $= \sqrt{104976 + 186624}$ \\
            $= \sqrt{291600}$ \\
            $= 540$
            
            \vspace{0.5em}
            Therefore, the largest possible real part is $\boxed{540}$.
        \end{minipage}%
    }
    \caption{Solution generated by \ourmodel-Qwen-7B for an AIME2024 problem.}
    \label{fig:case_study_7b_5}
\end{figure}

%% file: acl_latex.bbl
\begin{thebibliography}{53}
\providecommand{\natexlab}[1]{#1}

\bibitem[{AIME-2024()}]{aimedata}
AIME-2024.
\newblock https://huggingface.co/datasets/ai-mo/aimo-validation-aime.

\bibitem[{Azerbayev et~al.(2024)Azerbayev, Schoelkopf, Paster, Santos, McAleer, Jiang, Deng, Biderman, and Welleck}]{azerbayev2024llemmaopenlanguagemodel}
Zhangir Azerbayev, Hailey Schoelkopf, Keiran Paster, Marco~Dos Santos, Stephen McAleer, Albert~Q. Jiang, Jia Deng, Stella Biderman, and Sean Welleck. 2024.
\newblock \href {https://arxiv.org/abs/2310.10631} {Llemma: An open language model for mathematics}.
\newblock \emph{Preprint}, arXiv:2310.10631.

\bibitem[{Cheng et~al.(2024)Cheng, Guan, Wu, and Yan}]{cheng2024mostbuildingplugandplayvisual}
Chuanqi Cheng, Jian Guan, Wei Wu, and Rui Yan. 2024.
\newblock \href {https://arxiv.org/abs/2406.19934} {From the least to the most: Building a plug-and-play visual reasoner via data synthesis}.
\newblock \emph{Preprint}, arXiv:2406.19934.

\bibitem[{Cobbe et~al.(2021)Cobbe, Kosaraju, Bavarian, Chen, Jun, Kaiser, Plappert, Tworek, Hilton, Nakano et~al.}]{cobbe2021training}
Karl Cobbe, Vineet Kosaraju, Mohammad Bavarian, Mark Chen, Heewoo Jun, Lukasz Kaiser, Matthias Plappert, Jerry Tworek, Jacob Hilton, Reiichiro Nakano, et~al. 2021.
\newblock Training verifiers to solve math word problems.
\newblock \emph{arXiv preprint arXiv:2110.14168}.

\bibitem[{Dubey et~al.(2024)Dubey, Jauhri, Pandey, Kadian, Al-Dahle, Letman, Mathur, Schelten, Yang, Fan et~al.}]{dubey2024llama}
Abhimanyu Dubey, Abhinav Jauhri, Abhinav Pandey, Abhishek Kadian, Ahmad Al-Dahle, Aiesha Letman, Akhil Mathur, Alan Schelten, Amy Yang, Angela Fan, et~al. 2024.
\newblock The llama 3 herd of models.
\newblock \emph{arXiv preprint arXiv:2407.21783}.

\bibitem[{Gao et~al.(2023)Gao, Madaan, Zhou, Alon, Liu, Yang, Callan, and Neubig}]{gao2023pal}
Luyu Gao, Aman Madaan, Shuyan Zhou, Uri Alon, Pengfei Liu, Yiming Yang, Jamie Callan, and Graham Neubig. 2023.
\newblock Pal: Program-aided language models.
\newblock In \emph{International Conference on Machine Learning}, pages 10764--10799. PMLR.

\bibitem[{Ge et~al.(2024)Ge, Chan, Wang, Yu, Mi, and Yu}]{ge2024scalingsyntheticdatacreation}
Tao Ge, Xin Chan, Xiaoyang Wang, Dian Yu, Haitao Mi, and Dong Yu. 2024.
\newblock \href {https://arxiv.org/abs/2406.20094} {Scaling synthetic data creation with 1,000,000,000 personas}.
\newblock \emph{Preprint}, arXiv:2406.20094.

\bibitem[{Guan et~al.(2025)Guan, Zhang, Liu, Shang, Sun, Zhu, Yang, and Yang}]{guan2025rstar}
Xinyu Guan, Li~Lyna Zhang, Yifei Liu, Ning Shang, Youran Sun, Yi~Zhu, Fan Yang, and Mao Yang. 2025.
\newblock rstar-math: Small llms can master math reasoning with self-evolved deep thinking.
\newblock \emph{arXiv preprint arXiv:2501.04519}.

\bibitem[{Guo et~al.(2025)Guo, Yang, Zhang, Song, Zhang, Xu, Zhu, Ma, Wang, Bi et~al.}]{guo2025deepseek}
Daya Guo, Dejian Yang, Haowei Zhang, Junxiao Song, Ruoyu Zhang, Runxin Xu, Qihao Zhu, Shirong Ma, Peiyi Wang, Xiao Bi, et~al. 2025.
\newblock Deepseek-r1: Incentivizing reasoning capability in llms via reinforcement learning.
\newblock \emph{arXiv preprint arXiv:2501.12948}.

\bibitem[{Huang et~al.(2024)Huang, Liu, Gong, Gou, Shen, Duan, and Chen}]{huang2024key}
Yiming Huang, Xiao Liu, Yeyun Gong, Zhibin Gou, Yelong Shen, Nan Duan, and Weizhu Chen. 2024.
\newblock Key-point-driven data synthesis with its enhancement on mathematical reasoning.
\newblock \emph{arXiv preprint arXiv:2403.02333}.

\bibitem[{Imani et~al.(2023)Imani, Du, and Shrivastava}]{imani2023mathprompter}
Shima Imani, Liang Du, and Harsh Shrivastava. 2023.
\newblock Mathprompter: Mathematical reasoning using large language models.
\newblock In \emph{Proceedings of the 61st Annual Meeting of the Association for Computational Linguistics (Volume 5: Industry Track)}, pages 37--42.

\bibitem[{Jaech et~al.(2024)Jaech, Kalai, Lerer, Richardson, El-Kishky, Low, Helyar, Madry, Beutel, Carney et~al.}]{jaech2024openai}
Aaron Jaech, Adam Kalai, Adam Lerer, Adam Richardson, Ahmed El-Kishky, Aiden Low, Alec Helyar, Aleksander Madry, Alex Beutel, Alex Carney, et~al. 2024.
\newblock Openai o1 system card.
\newblock \emph{arXiv preprint arXiv:2412.16720}.

\bibitem[{Kambhampati(2024)}]{Kambhampati_2024}
Subbarao Kambhampati. 2024.
\newblock \href {https://doi.org/10.1111/nyas.15125} {Can large language models reason and plan?}
\newblock \emph{Annals of the New York Academy of Sciences}, 1534(1):15–18.

\bibitem[{Li et~al.(2024{\natexlab{a}})Li, Wang, Hu, Wei, Zheng, Hu, Zhang, and Peng}]{li2024common7blanguagemodels}
Chen Li, Weiqi Wang, Jingcheng Hu, Yixuan Wei, Nanning Zheng, Han Hu, Zheng Zhang, and Houwen Peng. 2024{\natexlab{a}}.
\newblock \href {https://arxiv.org/abs/2403.04706} {Common 7b language models already possess strong math capabilities}.
\newblock \emph{Preprint}, arXiv:2403.04706.

\bibitem[{Li et~al.(2024{\natexlab{b}})Li, Beeching, Tunstall, Lipkin, Soletskyi, Huang, Rasul, Yu, Jiang, Shen et~al.}]{li2024numinamath}
Jia Li, Edward Beeching, Lewis Tunstall, Ben Lipkin, Roman Soletskyi, Shengyi Huang, Kashif Rasul, Longhui Yu, Albert~Q Jiang, Ziju Shen, et~al. 2024{\natexlab{b}}.
\newblock Numinamath: The largest public dataset in ai4maths with 860k pairs of competition math problems and solutions.
\newblock \emph{Hugging Face repository}, 13:9.

\bibitem[{Li et~al.(2024{\natexlab{c}})Li, Cui, Zhao, Kong, and Bi}]{li2024gsm}
Qintong Li, Leyang Cui, Xueliang Zhao, Lingpeng Kong, and Wei Bi. 2024{\natexlab{c}}.
\newblock Gsm-plus: A comprehensive benchmark for evaluating the robustness of llms as mathematical problem solvers.
\newblock \emph{arXiv preprint arXiv:2402.19255}.

\bibitem[{Li et~al.(2024{\natexlab{d}})Li, Yu, Zhou, Schick, Levy, Zettlemoyer, Weston, and Lewis}]{li2024selfalignmentinstructionbacktranslation}
Xian Li, Ping Yu, Chunting Zhou, Timo Schick, Omer Levy, Luke Zettlemoyer, Jason Weston, and Mike Lewis. 2024{\natexlab{d}}.
\newblock \href {https://arxiv.org/abs/2308.06259} {Self-alignment with instruction backtranslation}.
\newblock \emph{Preprint}, arXiv:2308.06259.

\bibitem[{Lightman et~al.(2023)Lightman, Kosaraju, Burda, Edwards, Baker, Lee, Leike, Schulman, Sutskever, and Cobbe}]{lightman2023let}
Hunter Lightman, Vineet Kosaraju, Yura Burda, Harri Edwards, Bowen Baker, Teddy Lee, Jan Leike, John Schulman, Ilya Sutskever, and Karl Cobbe. 2023.
\newblock Let's verify step by step.
\newblock \emph{arXiv preprint arXiv:2305.20050}.

\bibitem[{Luo et~al.(2023)Luo, Sun, Xu, Zhao, Lou, Tao, Geng, Lin, Chen, and Zhang}]{luo2023wizardmath}
Haipeng Luo, Qingfeng Sun, Can Xu, Pu~Zhao, Jianguang Lou, Chongyang Tao, Xiubo Geng, Qingwei Lin, Shifeng Chen, and Dongmei Zhang. 2023.
\newblock Wizardmath: Empowering mathematical reasoning for large language models via reinforced evol-instruct.
\newblock \emph{arXiv preprint arXiv:2308.09583}.

\bibitem[{Mirzadeh et~al.(2024)Mirzadeh, Alizadeh, Shahrokhi, Tuzel, Bengio, and Farajtabar}]{mirzadeh2024gsmsymbolicunderstandinglimitationsmathematical}
Iman Mirzadeh, Keivan Alizadeh, Hooman Shahrokhi, Oncel Tuzel, Samy Bengio, and Mehrdad Farajtabar. 2024.
\newblock \href {https://arxiv.org/abs/2410.05229} {Gsm-symbolic: Understanding the limitations of mathematical reasoning in large language models}.
\newblock \emph{Preprint}, arXiv:2410.05229.

\bibitem[{Mitra et~al.(2024)Mitra, Khanpour, Rosset, and Awadallah}]{mitra2024orca}
Arindam Mitra, Hamed Khanpour, Corby Rosset, and Ahmed Awadallah. 2024.
\newblock Orca-math: Unlocking the potential of slms in grade school math.
\newblock \emph{arXiv preprint arXiv:2402.14830}.

\bibitem[{Muennighoff et~al.(2025)Muennighoff, Yang, Shi, Li, Fei-Fei, Hajishirzi, Zettlemoyer, Liang, Candès, and Hashimoto}]{muennighoff2025s1simpletesttimescaling}
Niklas Muennighoff, Zitong Yang, Weijia Shi, Xiang~Lisa Li, Li~Fei-Fei, Hannaneh Hajishirzi, Luke Zettlemoyer, Percy Liang, Emmanuel Candès, and Tatsunori Hashimoto. 2025.
\newblock \href {https://arxiv.org/abs/2501.19393} {s1: Simple test-time scaling}.
\newblock \emph{Preprint}, arXiv:2501.19393.

\bibitem[{OpenAI(2024{\natexlab{a}})}]{openai2024o1preview}
OpenAI. 2024{\natexlab{a}}.
\newblock Introducing openai o1-preview.
\newblock \url{https://openai.com/index/introducing-openai-o1-preview/}.

\bibitem[{OpenAI(2024{\natexlab{b}})}]{openai-o1}
OpenAI. 2024{\natexlab{b}}.
\newblock \href {https://openai.com/index/learning-to-reason-with-llms/} {Learning to reason with llms, september 2024}.

\bibitem[{Paster et~al.(2023)Paster, Santos, Azerbayev, and Ba}]{paster2023openwebmathopendatasethighquality}
Keiran Paster, Marco~Dos Santos, Zhangir Azerbayev, and Jimmy Ba. 2023.
\newblock \href {https://arxiv.org/abs/2310.06786} {Openwebmath: An open dataset of high-quality mathematical web text}.
\newblock \emph{Preprint}, arXiv:2310.06786.

\bibitem[{Press et~al.(2023)Press, Zhang, Min, Schmidt, Smith, and Lewis}]{press2023measuring}
Ofir Press, Muru Zhang, Sewon Min, Ludwig Schmidt, Noah~A Smith, and Mike Lewis. 2023.
\newblock Measuring and narrowing the compositionality gap in language models.
\newblock In \emph{Findings of the Association for Computational Linguistics: EMNLP 2023}, pages 5687--5711.

\bibitem[{Schick et~al.(2024)Schick, Dwivedi-Yu, Dess{\`\i}, Raileanu, Lomeli, Hambro, Zettlemoyer, Cancedda, and Scialom}]{schick2024toolformer}
Timo Schick, Jane Dwivedi-Yu, Roberto Dess{\`\i}, Roberta Raileanu, Maria Lomeli, Eric Hambro, Luke Zettlemoyer, Nicola Cancedda, and Thomas Scialom. 2024.
\newblock Toolformer: Language models can teach themselves to use tools.
\newblock \emph{Advances in Neural Information Processing Systems}, 36.

\bibitem[{Shao et~al.(2024)Shao, Wang, Zhu, Xu, Song, Zhang, Li, Wu, and Guo}]{shao2024deepseekmath}
Zhihong Shao, Peiyi Wang, Qihao Zhu, Runxin Xu, Junxiao Song, Mingchuan Zhang, YK~Li, Yu~Wu, and Daya Guo. 2024.
\newblock Deepseekmath: Pushing the limits of mathematical reasoning in open language models.
\newblock \emph{arXiv preprint arXiv:2402.03300}.

\bibitem[{Snell et~al.(2024)Snell, Lee, Xu, and Kumar}]{snell2024scaling}
Charlie Snell, Jaehoon Lee, Kelvin Xu, and Aviral Kumar. 2024.
\newblock Scaling llm test-time compute optimally can be more effective than scaling model parameters.
\newblock \emph{arXiv preprint arXiv:2408.03314}.

\bibitem[{Tang et~al.(2024)Tang, Zhang, Wang, and Wei}]{tang2024mathscale}
Zhengyang Tang, Xingxing Zhang, Benyou Wang, and Furu Wei. 2024.
\newblock Mathscale: Scaling instruction tuning for mathematical reasoning.
\newblock \emph{arXiv preprint arXiv:2403.02884}.

\bibitem[{Team(2024)}]{qwq}
Qwen Team. 2024.
\newblock \href {https://qwenlm.github.io/blog/qwq-32b-preview/} {Qwq: Reflect deeply on the boundaries of the unknown}.

\bibitem[{Toshniwal et~al.(2024)Toshniwal, Du, Moshkov, Kisacanin, Ayrapetyan, and Gitman}]{toshniwal2024openmathinstruct}
Shubham Toshniwal, Wei Du, Ivan Moshkov, Branislav Kisacanin, Alexan Ayrapetyan, and Igor Gitman. 2024.
\newblock Openmathinstruct-2: Accelerating ai for math with massive open-source instruction data.
\newblock \emph{arXiv preprint arXiv:2410.01560}.

\bibitem[{Wang et~al.(2022)Wang, Wei, Schuurmans, Le, Chi, Narang, Chowdhery, and Zhou}]{wang2022self}
Xuezhi Wang, Jason Wei, Dale Schuurmans, Quoc Le, Ed~Chi, Sharan Narang, Aakanksha Chowdhery, and Denny Zhou. 2022.
\newblock Self-consistency improves chain of thought reasoning in language models.
\newblock \emph{arXiv preprint arXiv:2203.11171}.

\bibitem[{Wang et~al.(2023)Wang, Kordi, Mishra, Liu, Smith, Khashabi, and Hajishirzi}]{wang2023selfinstructaligninglanguagemodels}
Yizhong Wang, Yeganeh Kordi, Swaroop Mishra, Alisa Liu, Noah~A. Smith, Daniel Khashabi, and Hannaneh Hajishirzi. 2023.
\newblock \href {https://arxiv.org/abs/2212.10560} {Self-instruct: Aligning language models with self-generated instructions}.
\newblock \emph{Preprint}, arXiv:2212.10560.

\bibitem[{Wang et~al.(2024)Wang, Li, Xia, and Liu}]{wang2024mathpilebilliontokenscalepretrainingcorpus}
Zengzhi Wang, Xuefeng Li, Rui Xia, and Pengfei Liu. 2024.
\newblock \href {https://arxiv.org/abs/2312.17120} {Mathpile: A billion-token-scale pretraining corpus for math}.
\newblock \emph{Preprint}, arXiv:2312.17120.

\bibitem[{Wei et~al.(2022{\natexlab{a}})Wei, Wang, Schuurmans, Bosma, Xia, Chi, Le, Zhou et~al.}]{wei2022chain}
Jason Wei, Xuezhi Wang, Dale Schuurmans, Maarten Bosma, Fei Xia, Ed~Chi, Quoc~V Le, Denny Zhou, et~al. 2022{\natexlab{a}}.
\newblock Chain-of-thought prompting elicits reasoning in large language models.
\newblock \emph{Advances in neural information processing systems}, 35:24824--24837.

\bibitem[{Wei et~al.(2022{\natexlab{b}})Wei, Wang, Schuurmans, Bosma, Xia, Chi, Le, Zhou et~al.}]{cot}
Jason Wei, Xuezhi Wang, Dale Schuurmans, Maarten Bosma, Fei Xia, Ed~Chi, Quoc~V Le, Denny Zhou, et~al. 2022{\natexlab{b}}.
\newblock Chain-of-thought prompting elicits reasoning in large language models.
\newblock \emph{Advances in neural information processing systems}, 35:24824--24837.

\bibitem[{Wen et~al.(2025)Wen, Guan, Wang, Wu, and Huang}]{wen2025unlocking}
Jiaxin Wen, Jian Guan, Hongning Wang, Wei Wu, and Minlie Huang. 2025.
\newblock \href {https://openreview.net/forum?id=dCPF1wlqj8} {Unlocking reasoning potential in large language models by scaling code-form planning}.
\newblock In \emph{The Thirteenth International Conference on Learning Representations}.

\bibitem[{Wu et~al.(2024)Wu, Sun, Li, Welleck, and Yang}]{wu2024inference}
Yangzhen Wu, Zhiqing Sun, Shanda Li, Sean Welleck, and Yiming Yang. 2024.
\newblock Inference scaling laws: An empirical analysis of compute-optimal inference for problem-solving with language models.
\newblock \emph{arXiv preprint arXiv:2408.00724}.

\bibitem[{Xu et~al.(2023)Xu, Sun, Zheng, Geng, Zhao, Feng, Tao, and Jiang}]{xu2023wizardlm}
Can Xu, Qingfeng Sun, Kai Zheng, Xiubo Geng, Pu~Zhao, Jiazhan Feng, Chongyang Tao, and Daxin Jiang. 2023.
\newblock Wizardlm: Empowering large language models to follow complex instructions.
\newblock \emph{arXiv preprint arXiv:2304.12244}.

\bibitem[{Xu et~al.(2024)Xu, Jiang, Niu, Deng, Poovendran, Choi, and Lin}]{xu2024magpiealignmentdatasynthesis}
Zhangchen Xu, Fengqing Jiang, Luyao Niu, Yuntian Deng, Radha Poovendran, Yejin Choi, and Bill~Yuchen Lin. 2024.
\newblock \href {https://arxiv.org/abs/2406.08464} {Magpie: Alignment data synthesis from scratch by prompting aligned llms with nothing}.
\newblock \emph{Preprint}, arXiv:2406.08464.

\bibitem[{Yang et~al.(2024{\natexlab{a}})Yang, Yang, Zhang, Hui, Zheng, Yu, Li, Liu, Huang, Wei et~al.}]{yang2024qwen25llm}
An~Yang, Baosong Yang, Beichen Zhang, Binyuan Hui, Bo~Zheng, Bowen Yu, Chengyuan Li, Dayiheng Liu, Fei Huang, Haoran Wei, et~al. 2024{\natexlab{a}}.
\newblock Qwen2. 5 technical report.
\newblock \emph{arXiv preprint arXiv:2412.15115}.

\bibitem[{Yang et~al.(2024{\natexlab{b}})Yang, Zhang, Hui, Gao, Yu, Li, Liu, Tu, Zhou, Lin et~al.}]{yang2024qwen2}
An~Yang, Beichen Zhang, Binyuan Hui, Bofei Gao, Bowen Yu, Chengpeng Li, Dayiheng Liu, Jianhong Tu, Jingren Zhou, Junyang Lin, et~al. 2024{\natexlab{b}}.
\newblock Qwen2. 5-math technical report: Toward mathematical expert model via self-improvement.
\newblock \emph{arXiv preprint arXiv:2409.12122}.

\bibitem[{Ye et~al.(2025)Ye, Huang, Xiao, Chern, Xia, and Liu}]{ye2025limoreasoning}
Yixin Ye, Zhen Huang, Yang Xiao, Ethan Chern, Shijie Xia, and Pengfei Liu. 2025.
\newblock \href {https://arxiv.org/abs/2502.03387} {Limo: Less is more for reasoning}.
\newblock \emph{Preprint}, arXiv:2502.03387.

\bibitem[{Yu et~al.(2023)Yu, Jiang, Shi, Yu, Liu, Zhang, Kwok, Li, Weller, and Liu}]{yu2023metamath}
Longhui Yu, Weisen Jiang, Han Shi, Jincheng Yu, Zhengying Liu, Yu~Zhang, James~T Kwok, Zhenguo Li, Adrian Weller, and Weiyang Liu. 2023.
\newblock Metamath: Bootstrap your own mathematical questions for large language models.
\newblock \emph{arXiv preprint arXiv:2309.12284}.

\bibitem[{Yu et~al.(2024)Yu, Jiang, Shi, Yu, Liu, Zhang, Kwok, Li, Weller, and Liu}]{yu2024metamathbootstrapmathematicalquestions}
Longhui Yu, Weisen Jiang, Han Shi, Jincheng Yu, Zhengying Liu, Yu~Zhang, James~T. Kwok, Zhenguo Li, Adrian Weller, and Weiyang Liu. 2024.
\newblock \href {https://arxiv.org/abs/2309.12284} {Metamath: Bootstrap your own mathematical questions for large language models}.
\newblock \emph{Preprint}, arXiv:2309.12284.

\bibitem[{Yue et~al.(2023)Yue, Qu, Zhang, Fu, Huang, Sun, Su, and Chen}]{yue2023mammothbuildingmathgeneralist}
Xiang Yue, Xingwei Qu, Ge~Zhang, Yao Fu, Wenhao Huang, Huan Sun, Yu~Su, and Wenhu Chen. 2023.
\newblock \href {https://arxiv.org/abs/2309.05653} {Mammoth: Building math generalist models through hybrid instruction tuning}.
\newblock \emph{Preprint}, arXiv:2309.05653.

\bibitem[{Yue et~al.(2024)Yue, Zheng, Zhang, and Chen}]{yue2024mammoth2scalinginstructionsweb}
Xiang Yue, Tuney Zheng, Ge~Zhang, and Wenhu Chen. 2024.
\newblock \href {https://arxiv.org/abs/2405.03548} {Mammoth2: Scaling instructions from the web}.
\newblock \emph{Preprint}, arXiv:2405.03548.

\bibitem[{Zhang et~al.(2024)Zhang, Da, Lee, Robinson, Wu, Song, Zhao, Raja, Zhuang, Slack, Lyu, Hendryx, Kaplan, Lunati, and Yue}]{zhang2024carefulexaminationlargelanguage}
Hugh Zhang, Jeff Da, Dean Lee, Vaughn Robinson, Catherine Wu, Will Song, Tiffany Zhao, Pranav Raja, Charlotte Zhuang, Dylan Slack, Qin Lyu, Sean Hendryx, Russell Kaplan, Michele Lunati, and Summer Yue. 2024.
\newblock \href {https://arxiv.org/abs/2405.00332} {A careful examination of large language model performance on grade school arithmetic}.
\newblock \emph{Preprint}, arXiv:2405.00332.

\bibitem[{Zhao et~al.(2024{\natexlab{a}})Zhao, Huang, Bi, and Kong}]{zhao2024sego}
Xueliang Zhao, Xinting Huang, Wei Bi, and Lingpeng Kong. 2024{\natexlab{a}}.
\newblock Sego: Sequential subgoal optimization for mathematical problem-solving.
\newblock In \emph{The 62nd Annual Meeting of the Association for Computational Linguistics (11/08/2024-16/08/2024, Bangkok, Thailand)}.

\bibitem[{Zhao et~al.(2024{\natexlab{b}})Zhao, Huang, Fu, Li, Gong, Liu, Bi, and Kong}]{zhao2024bba}
Xueliang Zhao, Xinting Huang, Tingchen Fu, Qintong Li, Shansan Gong, Lemao Liu, Wei Bi, and Lingpeng Kong. 2024{\natexlab{b}}.
\newblock Bba: Bi-modal behavioral alignment for reasoning with large vision-language models.
\newblock In \emph{ACL (Findings)}.

\bibitem[{Zhao et~al.(2024{\natexlab{c}})Zhao, Li, and Kong}]{zhao2024subgoal}
Xueliang Zhao, Wenda Li, and Lingpeng Kong. 2024{\natexlab{c}}.
\newblock Subgoal-based demonstration learning for formal theorem proving.
\newblock In \emph{Forty-first International Conference on Machine Learning}.

\bibitem[{Zhao et~al.(2024{\natexlab{d}})Zhao, Zheng, Bo, Hu, Thakker, and Kong}]{zhao2024subgoalxl}
Xueliang Zhao, Lin Zheng, Haige Bo, Changran Hu, Urmish Thakker, and Lingpeng Kong. 2024{\natexlab{d}}.
\newblock Subgoalxl: Subgoal-based expert learning for theorem proving.
\newblock \emph{arXiv preprint arXiv:2408.11172}.

\end{thebibliography}
